\documentclass[lettersize,journal]{IEEEtran}
\usepackage{amsmath,amsfonts}
\usepackage{algorithmic}
\usepackage{algorithm}
\usepackage{array}
\usepackage{textcomp}
\usepackage{stfloats}
\usepackage{url}
\usepackage{verbatim}
\usepackage{graphicx}
\usepackage{cite}
\usepackage{booktabs}
\usepackage{multirow}
\usepackage[caption=false,labelformat=simple]{subfig}
\usepackage{authblk}
\usepackage{threeparttable}
\usepackage{graphicx}

\begin{document}

\title{Multi-Scale Adaptive Graph Neural Network for Multivariate Time Series Forecasting}

\author{Ling Chen, Donghui Chen,  Zongjiang Shang, Binqing Wu, Cen Zheng, Bo Wen, and Wei Zhang
\thanks{{Ling Chen and Donghui Chen are co-first authors.} {Corresponding author: Ling Chen}.}      
\thanks{Ling Chen, Donghui Chen, Zongjiang Shang, and Binqing Wu are with the College of Computer Science and Technology, Zhejiang University, Hangzhou 310027, China (e-mail: \{lingchen, chendonghui, zongjiangshang, binqingwu\}@cs.zju.edu.cn).}
\thanks{Cen Zheng, Bo Wen, and Wei Zhang are with Alibaba Group, Hangzhou 311100, China (e-mail: \{mingyan.zc, wenbo.wb, zwei\}@alibaba-inc.com).}}


\def\mathbi#1{\textbf{\em #1}}
\maketitle

\begin{abstract}
Multivariate time series (MTS) forecasting plays an important role in the automation and optimization of intelligent applications. It is a challenging task, as we need to consider both complex intra-variable dependencies and inter-variable dependencies. Existing works only learn temporal patterns with the help of single inter-variable dependencies. However, there are multi-scale temporal patterns in many real-world MTS. Single inter-variable dependencies make the model prefer to learn one type of prominent and shared temporal patterns. In this paper, we propose a multi-scale adaptive graph neural network (MAGNN) to address the above issue. MAGNN exploits a multi-scale pyramid network to preserve the underlying temporal dependencies at different time scales. Since the inter-variable dependencies may be different under distinct time scales, an adaptive graph learning module is designed to infer the scale-specific inter-variable dependencies without pre-defined priors. Given the multi-scale feature representations and scale-specific inter-variable dependencies, a multi-scale temporal graph neural network is introduced to jointly model intra-variable dependencies and inter-variable dependencies. After that, we develop a scale-wise fusion module to effectively promote the collaboration across different time scales, and automatically capture the importance of contributed temporal patterns. Experiments on six real-world datasets demonstrate that MAGNN outperforms the state-of-the-art methods across various settings. 
\end{abstract}


\section{Introduction}
Multivariate time series (MTS) are ubiquitous in various real-world scenarios, e.g., the traffic flows in a city, the stock prices in a stock market, and the household power consumption in a city block \cite{ref1}. MTS forecasting, which aims at forecasting the future trends based on a group of historical observed time series, has been widely studied in recent years. It is of great importance in a wide range of applications, e.g., a better driving route can be planned in advance based on the forecasted traffic flows, and an investment strategy can be designed with the forecasting of the near-future stock market \cite{ref2,ref3,ref4,reftkde1}.

Making accurate MTS forecasting is a challenging task, as both intra-variable dependencies (i.e., the temporal dependencies within one time series) and inter-variable dependencies (i.e., the forecasting values of a single variable are affected by other variables) need to be considered jointly. To solve this problem, traditional methods \cite{ref5,ref6,ref7}, e.g., vector auto-regression (VAR), temporal regularized matrix factorization (TRMF), vector auto-regression moving average (VARMA), and gaussian process (GP), often rely on the strict stationary assumption and cannot capture the non-linear dependencies among variables. Deep neural networks have shown superiority on modeling non-stationary and non-linear dependencies. Particularly, two variants of recurrent neural network (RNNs) \cite{ref8}, namely the long-short term memory (LSTM) and the gated recurrent unit (GRU), and temporal convolutional networks (TCNs) \cite{ref9} have significantly achieved impressive performance in time series modeling. To capture both long-term and short-term temporal dependencies, existing works \cite{ref3,refinsert1,ref10, ref11,refinsert2} introduce several strategies, e.g., skip-connection, attention mechanism, and memory-based network. These works focus on modeling temporal dependencies, and process the MTS input as vectors and assume that the forecasting values of a single variable are affected by all other variables, which is unreasonable and hard to meet in realistic applications. For example, the traffic flows of a street are largely affected by its neighboring streets, while the impact from distant streets is relatively small. Thus, it is crucial to model the pairwise inter-variable dependencies explicitly.

Graph is an abstract data type representing relations between nodes. Graph neural networks (GNNs) \cite{ref12, ref13}, which can effectively capture nodes’ high-level representations while exploiting pairwise dependencies, have been considered as a promising way to handle graph data. MTS forecasting can be considered from the perspective of graph modeling. The variables in MTS can be regarded as the nodes in a graph, while the pairwise inter-variable dependencies as edges. Recently, several works \cite{ref14,ref15,ref16} exploit GNNs to model MTS taking advantage of the rich structural information (i.e., featured nodes and weighted edges) of a graph. These works stack GNN and temporal convolution modules to learn temporal patterns, and have achieved promising results. Nevertheless, there are still two important aspects neglected in above works.

\begin{figure}[!t]
\centering
\includegraphics[width=3.5in]{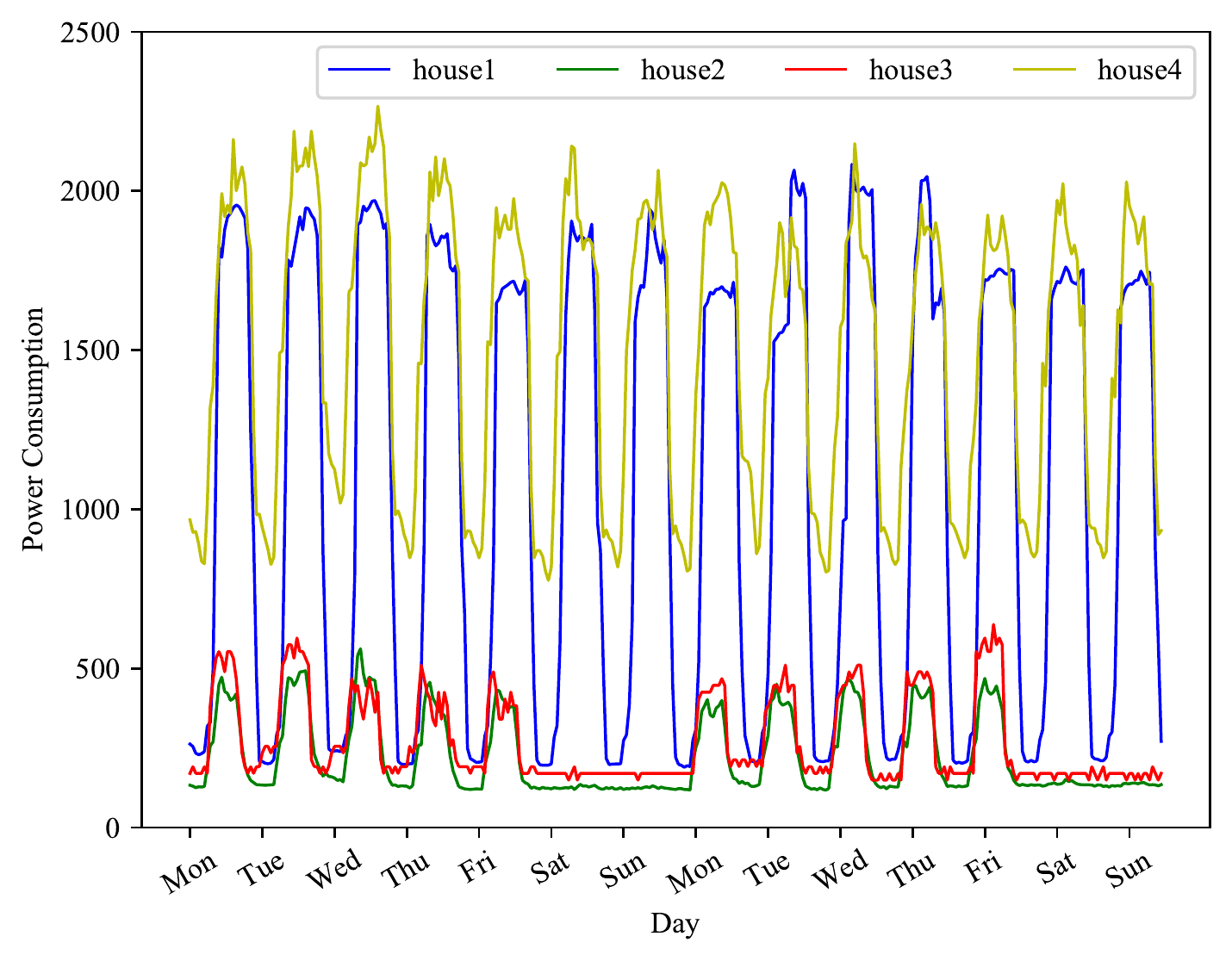}
\caption{The power consumptions of 4 households within two weeks (from Monday 00:00 to Sunday 24:00). Households 1 and 4 have both daily and weekly repeating patterns, while households 2 and 3 have weekly repeating patterns.}
\label{Figure_1}
\end{figure}

First, existing works only consider temporal dependencies on a single time scale, which may not properly reflect the variations in many real-world scenarios. In fact, the temporal patterns hidden in real-world MTS are much more complicated, including daily, weekly, monthly, and other specific periodic patterns. For example, Fig. \ref{Figure_1} shows the power consumptions of 4 households within two weeks. There exists a mixture of short-term and long-term repeating patterns (i.e., daily and weekly). These multi-scale temporal patterns provide abundant information to model MTS. Furthermore, if the temporal patterns are learned from different time scales separately, and are then straightforwardly concatenated to obtain the final representation, the model is failed to capture cross-scale relationships and cannot focus on contributed temporal patterns. Thus, an accurate MTS forecasting model should learn a feature representation that can comprehensively reflect all kinds of multi-scale temporal patterns.

Second, existing works learn a shared adjacent matrix to represent the rich inter-variable dependencies, which makes the models be biased to learn one type of prominent and shared temporal patterns. In fact, different kinds of temporal patterns are often affected by different inter-variable dependencies, and we should distinguish the inter-variable dependencies when modeling distinct temporal patterns. For example, when modeling the short-term patterns of the power consumptions of a household, it might be essential to pay more attention to the power consumptions of its neighbors. Because the dynamics of short-term patterns are often affected by a common event, e.g., a transmission line fault decreases the power consumptions of a street block, and a sudden cold weather increases the power consumptions. When modeling the long-term patterns of the power consumptions of a household, it might be essential to pay more attention to the households that have similar living habits, e.g., working and sleeping hours, as these households would have similar daily and weekly temporal patterns. Therefore, the complicated inter-variable dependencies need to be fully considered when modeling these multi-scale temporal patterns.

In this paper, we propose a general framework termed Multi-scale Adaptive Graph Neural Network (MAGNN) for MTS forecasting to address above issues. Specifically, we introduce a multi-scale pyramid network to decompose the time series with different time scales in a hierarchical way. Then, an adaptive graph learning module is designed to automatically infer the scale-specific graph structures in the end-to-end framework, which can fully explore the abundant and implicit inter-variable dependencies under different time scales. After that, a multi-scale temporal graph neural network is incorporated into the framework to model intra-variable dependencies and inter-variable dependencies at each time scale. Finally, a scale-wise fusion module is designed to automatically consider the importance of scale-specific representations and capture the cross-scale correlations. In summary, our contributions are as follows:

\begin{itemize}
\item{Propose MAGNN, which learns a temporal representation that can comprehensively reflect both multi-scale temporal patterns and the scale-specific inter-variable dependencies.}
\item{Design an adaptive graph learning module to explore the abundant and implicit inter-variable dependencies under different time scales, and a scale-wise fusion module to promote the collaboration across these scale-specific temporal representations and automatically capture the importance of contributed temporal patterns.}
\item{Conduct extensive experiments on six real-world MTS benchmark datasets. The experiment results demonstrate that the performance of our method is better than that of the state-of-the-art methods.}
\end{itemize}

The remainder of this paper is organized as follows: Section \uppercase\expandafter{\romannumeral2} and Section \uppercase\expandafter{\romannumeral3} give a survey of related work and preliminaries. Section \uppercase\expandafter{\romannumeral4} describes the proposed MAGNN method. Section \uppercase\expandafter{\romannumeral5} presents the experimental results and Section \uppercase\expandafter{\romannumeral6} concludes the paper.

\section{Related Work}
We briefly review the related work from two aspects: the MTS forecasting and graph learning for MTS.
\subsection{MTS Forecasting}
The problem of time series forecasting has been studied for decades. One of the most prominent traditional methods used for time series forecasting is the auto-regressive integrated moving average (ARIMA) model, because of its statistical properties and the flexibility on integrating several linear models, including auto-regression (AR), moving average, and auto-regressive moving average. However, limited by the high computational complexity, ARIMA is infeasible to model MTS. Vector auto-regression (VAR) and vector auto-regression moving average (VARMA) are the extension of AR and ARIMA, respectively, that can model MTS. Gaussian process (GP) \cite{ref5} is a Bayesian method to model distributions over a continuous domain of functions. GP can be used as a prior over the function space in Bayesian inference and has been applied to MTS forecasting. However, these works often rely on the strict stationary assumption and cannot capture the non-linear dependencies among variables. 

Recently, deep learning-based methods have shown superior capability on capturing non-stationary and non-linear dependencies. Most of existing works rely on LSTM and GRU to capture temporal dependencies\cite{ref35}. Compared with RNN-based approaches, dilated 1D convolutions \cite{ref15, ref36} are able to handle long-range sequences. However, the dilation rates of dilated 1D convolutions may cause the loss of local information, which brings in negative effects on modeling short-term dependencies. Some other efforts exploit TCNs and self-attention mechanism \cite{ref17} to model long time series efficiently. To capture both long-term and short-term temporal dependencies, LSTNet \cite{ref3} introduces the convolutional neural network to capture short-term temporal dependencies, and a recurrent-skip layer that can exploit the long-term periodic property hidden in time series. TPA-LSTM \cite{ref11} utilizes an attention mechanism, which enables the model to extract important temporal patterns and focus on different time steps for different variables. MTNet \cite{ref10} exploits the memory component and attention mechanism to effectively capture long-term temporal dependencies and periodic patterns. However, these works assume that each variable affects all other variables equally, which is unreasonable and hard to meet in realistic applications.

\subsection{Graph Learning for MTS}
Graph neural networks (GNNs) \cite{ref12, ref13}, which can model the interaction between nodes through weighted edges, have received increasing attention. Recently, there are many works using GNNs to capture inter-variable dependencies in the area of MTS modeling. One of the challenges of the GNNs-based MTS forecasting is to obtain a well-defined graph structure as the inter-variable dependencies. To solve this problem, existing methods can be roughly divided into three major categories: prior-knowledge-based, rule-based, and learning-based methods.

Prior-knowledge-based methods \cite{ref18,ref19,ref20,ref21} often exploit the extra information (e.g., road networks, physical structures, and extra feature matrices) in their specific scenarios. For example, in traffic flow forecasting \cite{ref18, ref20}, the graph structure can be constructed by the connections of road networks. If there is a connected road between two nodes, an edge is constructed in the graph structure, as the traffic flow at the upstream node will affect the traffic flow at the downstream node. In skeleton-based action recognition \cite{ref21}, the graph structure can be constructed by the physical structure of the human body, e.g., the multiple joints on the same arm are linked by the human skeleton, and edges can be constructed between these joints. In the ride-hailing demand forecasting \cite{ref19}, multiple different graph structures are constructed from different views: the proximity of spatial distance, the connection of urban road network, and the similarity of region functionality. However, these methods require domain knowledge to design a graph structure, which is difficult to transfer between different scenarios.

Rule-based methods \cite{ref16, ref22,ref23,ref24}, as non-parametric methods, provide a data-driven manner to construct the graph structure. These methods usually include causal discovery (e.g., Granger causality and additive noise model) \cite{ref22, ref24}, entropy-based methods (e.g., transfer entropy and relative entropy) \cite{ref16}, similarity-based methods (e.g., Pearson correlation, mutual information, DTW distance, and edit distance) \cite{ref23}. For example, Huang et al. \cite{ref24} used Granger causality to construct a causal graph. Xu et al. \cite{ref16} calculated the pairwise transfer entropy between variables, which is regarded as the adjacency matrix of the graph structure. He et al. \cite{ref23} exploited dynamic time warping (DTW) algorithm, which is competent to capture the pattern similarities between two time series. However, these methods are non-parameterized methods and have limited flexibility, which can only learn a kind of specific inter-variable dependency.

Learning-based methods \cite{ref14, ref15, reftkde2,ref25,ref26,ref27,ref28,ref29} introduce a parameterized module to learn pairwise inter-variable dependencies automatically. Kipf et al. \cite{ref26} first introduced a neural relational inference model, which uses the original time series as input and exploits the variational inference to learn a graph structure. Subsequently, Webb et al. \cite{ref28} proposed a decomposition-based neural relational inference model to learn multiple types of graph structure. Graber et al. \cite{ref25} proposed a neural relational inference model that achieves different graph structures at each time step. The attention-based learning methods use the attention mechanism to learn the pairwise inter-variable dependencies. For traffic flow forecasting, Tang et al. \cite{ref27} used a graph attention module to learn graph structure. Zheng et al. \cite{ref29} used spatial attention mechanism to learn the correlation of traffic flow at different nodes. In addition, several works achieve this more directly, i.e., randomly intializing the representation of each node, and calculating the pairwise similarity of these nodes. The representations of the nodes can be optimized to obtain the most suitable value for the current data distribution. For example, Wu et al. \cite{ref15} exploited a graph learning module to learn inter-variable dependencies, and modelled MTS using the GNNs and dilated convolution networks. Bai et al. \cite{ref14} introduced a data adaptive graph generation module to infer the inter-variable dependencies and a node adaptive parameter learning module to capture node-specific features. However, existing works only learn single inter-variable dependencies, making the models biased to learn one type of prominent and shared temporal patterns among MTS.

\section{Preliminaries}
\subsection{Problem Formulation}
{\bf{Problem Statement.}} In this paper, we focus on MTS forecasting. Formally, given a sequence of observed time series signals $\boldsymbol{X}=\{\boldsymbol{x}_1,\boldsymbol{x}_2,...,\boldsymbol{x}_t,...,\boldsymbol{x}_T \}$, where $\boldsymbol{x}_t\in\mathbb{R}^{N\times 1}$ denotes the values at time step $t$, $N$ is the variable dimension, and $\boldsymbol{x}_{t,i}$ denotes the value of the $i^{\text{th}}$ variable at time step $t$, MTS forecasting aims at forecasting the future values $\boldsymbol{\widehat{x}}_{T+h}\in\mathbb{R}^{N\times 1}$ at time step $T+h$, where $h$ denotes the look-ahead horizon. The problem can be formulated as:  
\begin{equation}
\label{1}
\boldsymbol{\widehat{x}}_{T+h}=\mathcal{F}\left(\boldsymbol{x}_1,\boldsymbol{x}_2,...,\boldsymbol{x}_T;\theta\right),
\end{equation}
where $\mathcal{F}$ is the mapping function and $\theta$ denotes all learnable parameters.

Then, we give several definitions regarding MTS forecasting.

{\bf{Definition 1.}} $\emph{MTS to Graph.}$ A graph is defined as ${G}=\left({\boldsymbol{V}},\boldsymbol{E}\right)$, where ${\boldsymbol{V}}$ denotes the node set and $\left|{\boldsymbol{V}}\right|=N$. $\boldsymbol{E}$ is the edge set. Given the MTS $\boldsymbol{X}\in\mathbb{R}^{N{\times}T}$, the $i^{\text{th}}$ variable is regarded as the $i^{\text{th}}$ node $v_i\in{\boldsymbol{V}}$, the values of $\{\boldsymbol{x}_{1,i},\boldsymbol{x}_{2,i},...,\boldsymbol{x}_{T,i}\}$ are the features of $v_i$, and each edge $(v_i,v_j)\in\boldsymbol{E}$ indicates there is an inter-variable dependency between $v_i$ and $v_j$.

{\bf{Definition 2.}} $\emph{Weighted Adjacency Matrix.}$ The weighted adjacency matrix $\boldsymbol{{A}}\in\mathbb{R}^{N\times N}$ of a graph is a type of mathematical representation to store the weights of the edges, where $\boldsymbol{{A}}_{i,j}>0$, if $\left(v_i,v_j\right)\in\boldsymbol{E}$, and $\boldsymbol{{A}}_{i,j}=0$, if $\left(v_i,v_j\right)\notin\boldsymbol{E}$.

To pure MTS data without any prior knowledge, the weighted adjacency matrices of multiple graphs need to be learned to represent the abundant and implicit inter-variable dependencies. Accordingly, the formulation of MTS forecasting can be modified as:

\begin{equation}
\label{2}
\boldsymbol{\widehat{x}}_{T+h}=\mathcal{F}\left(\boldsymbol{x}_1,\boldsymbol{x}_2,...,\boldsymbol{x}_T;\boldsymbol{G};\theta\right),
\end{equation}
where $\boldsymbol{G}=\{G^1,G^2,...,G^K\}$ represents the set of graphs that can be utilized by GNNs for MTS forecasting.

\subsection{Graph Neural Networks}
Graph neural networks (GNNs) \cite{ref12,ref13} are a type of deep neural network applied to graphs. Graphs can be irregular, a graph may have a variable size of unordered nodes, and nodes from a graph may have different numbers of neighbors. GNNs can be easy to compute in the graph domain, which can overcome the limitation of CNNs.

GNNs can be divided into two categories based on the implementation philosophy: spectral-based and spatial-based methods \cite{ref12}. Spectral-based methods define the graph convolution by introducing a filter from the perspective of graph signal processing. The graph convolution operation can be interpreted as removing noise from the graph signal. Space-based methods define the graph convolution through information propagation, which aggregates the representation of a central node and the representations of its neighbors to get the updated representation for the node.

We briefly describe the graph convolution operation applied in our method, which can be defined as:
\begin{equation}
\label{3}
\boldsymbol{x} *{ }_{G} \theta=\sigma(\theta(\widetilde{\boldsymbol{D}}^{-\frac{1}{2}} \widetilde{\boldsymbol{A}} \widetilde{\boldsymbol{D}}^{-\frac{1}{2}}) \boldsymbol{x}),
\end{equation}
where $G=\left(\boldsymbol{V},\boldsymbol{E},\boldsymbol{A}\right)$ is a graph with a weighted adjacency matrix, $\boldsymbol{x}$ is the representations of nodes, $\sigma$ is an activation function, $\theta$ is the learnable parameter matrix,  $\widetilde{\boldsymbol{A}}=\boldsymbol{I}_n+\boldsymbol{A}$ is the adjacency matrix with self-connection, $\widetilde{\boldsymbol{D}}$ is the diagonal degree matrix of $\widetilde{\boldsymbol{A}}$, and $\widetilde{\boldsymbol{D}}_{i i}=\sum_{j} \widetilde{\boldsymbol{A}}_{i j}$. By stacking the graph convolution operation multiple layers, we can aggregate the information of multi-order neighbors.

Multi-scale GNNs \cite{ref40,ref41,ref42}, named hierarchical GNNs alternatively, usually construct coarse-grained graphs based on the fine-grained graph hierarchically. MAGNN is concerned about scales in the temporal dimension, which is very different from general multi-scale GNNs that focus on scales in the spatial dimension. MAGNN introduces a multi-scale pyramid network to transform raw time series into feature representations from smaller scale to larger scale, on which it learns scale-specific graphs with the same size for each scale and utilizes basic GNNs as defined in Eq. 3 for each graph.

\section{METHODOLOGY}
\subsection{Framework}
\begin{figure*}[!t]
\centering
\includegraphics[width=7.3in]{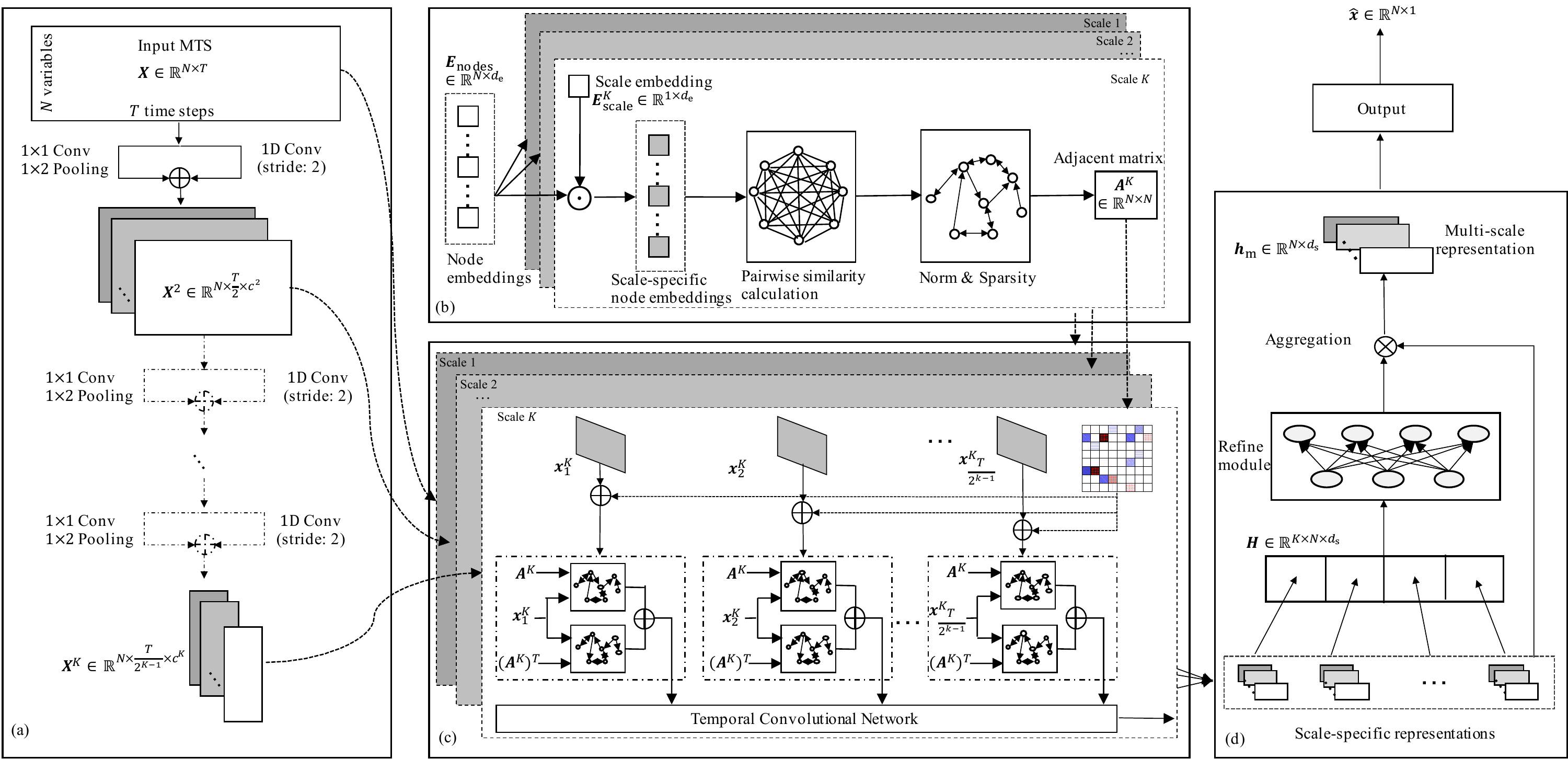}
\caption{The Multi-scale Adaptive Graph Neural Network (MAGNN) framework, which consists of four main parts: (a) The two parallel convolutional neural networks and point-wise additions at each layer transform feature representations from smaller scale to larger scale hierarchically. (b) An adaptive graph learning module takes node embeddings and scale embeddings as inputs and outputs the scale-specific adjacency matrices. (c) Each scale-specific feature representation and adjacency matrix are fed into a temporal GNN to obtain scale-specific representations. (d) Scale-specific representations are weighted fused to capture the contributed temporal patterns. The final multi-scale representation is fed into the output module including two convolutional neural networks to obtain the predicted values.}
\label{Figure_2}
\end{figure*}

Fig. \ref{Figure_2} illustrates the framework of MAGNN, which consists of four main parts: a) a multi-scale pyramid network to preserve the underlying temporal hierarchy at different time scales; b) an adaptive graph learning module to automatically infer inter-variable dependencies; c) a multi-scale temporal graph neural network to capture all kinds of scale-specific temporal patterns; d) a scale-wise fusion module to effectively promote the collaboration across different time scales.

\subsection{Multi-Scale Pyramid Network}
A multi-scale pyramid network is designed to preserve the underlying temporal dependencies at different time scales. Following the pyramid structure, it applies multiple pyramid layers to hierarchically transform raw time series into feature representations from smaller scale to larger scale. Such multi-scale structure gives us the opportunity to observe raw time series in different time scales. Specifically, the smaller scale feature representations can retain more fine-grained details, while the larger scale feature representations can capture the slow-varying trends.

Multi-scale pyramid network generates multi-scale feature representations through pyramid layers. Each pyramid layer takes the outputs of a preceding pyramid layer as the input and generates the feature representations of a larger scale as the output. Specifically, given the input MTS $\boldsymbol{X}\in\mathbb{R}^{N\times{T}}$, the multi-scale pyramid network generates feature representations of $K$ scales, and the $k^{\text{th}}$ scale feature representation is denoted as $\boldsymbol{X}^{k} \in \mathbb{R}^{N \times \frac{T}{2^{k-1}} \times c^{k}}$, where $N$ is the variable dimension, $\frac{T}{2^{k-1}}$ is the sequence length in the $k^{\text{th}}$ scale, and $c^k$ is the channel size of the $k^{\text{th}}$ scale.

A pyramid layer takes convolutional neural networks to capture local patterns in the time dimension. Following the design philosophy of image processing, different pyramid layers employ different kernel sizes. The beginning convolution kernel has larger filter, and the size is slowly decreased at each pyramid layer, which can control the receptive field size and maintain the sequence characteristics of large scale time series. For example, the kernel sizes can be set as 1$\times${7}, 1$\times${6}, and $1\times{3}$ at each pyramid layer, and the stride size of convolution is set to 2 to increase the time scale. Formally,
\begin{equation}
\label{4}
\boldsymbol{X}_{\mathrm{rec}}^{k}={ReLU}(\boldsymbol{W}_{\mathrm{rec}}^{k} \otimes \boldsymbol{X}^{k-1}+\boldsymbol{b}_{\mathrm{rec}}^{k}),
\end{equation}
where $\otimes$ denotes convolution operator, $\boldsymbol{W}_{*}^{k}$ and $\boldsymbol{b}_{*}^{k}$ denote the convolution kernel and bias vector in the $k^{\text{th}}$ pyramid layer, respectively. However, different pyramid layers are expected to preserve the underlying temporal dependencies at different time scales. The flexibility of using only one convolutional neural network is limited, as the granularities of the temporal dependencies captured in the feature representations at two consecutive pyramid layers are highly sensitive to the hyperparameter settings (i.e., kernel size and stride size). To alleviate this issue, following the existing works in image processing \cite{ref30, ref31}, we introduce another convolutional neural network with  kernel size $1\times{1}$ and a $1\times{2}$ pooling layer, which is a parallel structure with the original convolutional neural network, formally,
\begin{equation}
\label{5}
\boldsymbol{X}_{\text{norm}}^{k}=Pooling\left(ReLU(\boldsymbol{W}_{\text {norm }}^{k} \otimes \boldsymbol{X}^{k-1}+\boldsymbol{b}_{\text {norm }}^{k})\right).
\end{equation}

Then, a point-wise addition is utilized to the outputs of these two convolutional neural networks at each scale:
\begin{equation}
\label{6}
\boldsymbol{X}^{k}=\boldsymbol{X}_{\text {rec}}^{k}+\boldsymbol{X}_{\text {norm}}^{k}.
\end{equation}

After that, the learned multi-scale feature representations are flexible and comprehensive to preserve various kinds of temporal dependencies. During the process of feature representation learning, to avoid the interaction between the variables of MTS, the convolutional operations are performed on the time dimension, and the variable dimension is fixed, i.e., the kernels are shared between the variable dimension at each pyramid layer.

\subsection{Adaptive Graph Learning}
The adaptive graph learning module automatically generates adjacency matrices to represent the inter-variable dependencies among MTS. Existing learning-based methods \cite{ref12,ref13,ref38} only learn a shared adjacency matrix, which is useful to learn the most prominent inter-variable dependencies among MTS in many problems, and can significantly reduce the number of parameters and avoid the overfitting problem. However, the inter-variable dependencies may be different under different time scales. The shared adjacency matrix makes the models biased to learn one type of prominent and shared temporal patterns. Therefore, it is essential to learn multiple scale-specific adjacency matrices.

However, directly learning a unique adjacent matrix for each scale will introduce too many parameters and make the model hard to train, especially when the number of nodes is large\cite{ref39}. To solve this problem, we propose an adaptive graph learning (AGL) module that has $K$ scale-specific layers. Inspired by the matrix factorization, AGL has two kinds of parameters: 1) node embeddings $\boldsymbol{E}_{\text{nodes}}\in\mathbb{R}^{N\times{d}_{\text{e}}}$ shared between all scales, where $d_{\text{e}}$ is the embedding dimension and $d_{\text{e}}\ll{N}$; 2) scale embeddings $\boldsymbol{E}_{\text{scale}}\in\mathbb{R}^{K\times{d}_{\text{e}}}$. For the $k^{\text{th}}$ scale-specific layer, scale-specific node embeddings $\boldsymbol{E}_{\text{spec}}^k$ are obtained by the point-wise multiplication of the $k^{\text{th}}$ scale embedding $\boldsymbol{E}^{k}_{\text{scale}}\in\mathbb{R}^{1\times{d}_{\text{e}}}$ and node embeddings  $\boldsymbol{E}_{\text{nodes}}$:
\begin{equation}
\label{7}
\boldsymbol{E}^{k}_{\text{spec}}=\boldsymbol{E}_{\text{nodes}}\odot\boldsymbol{E}_\text{scale}^k.
\end{equation}

\begin{figure}[!t]
\centering
\includegraphics[width=3.5in]{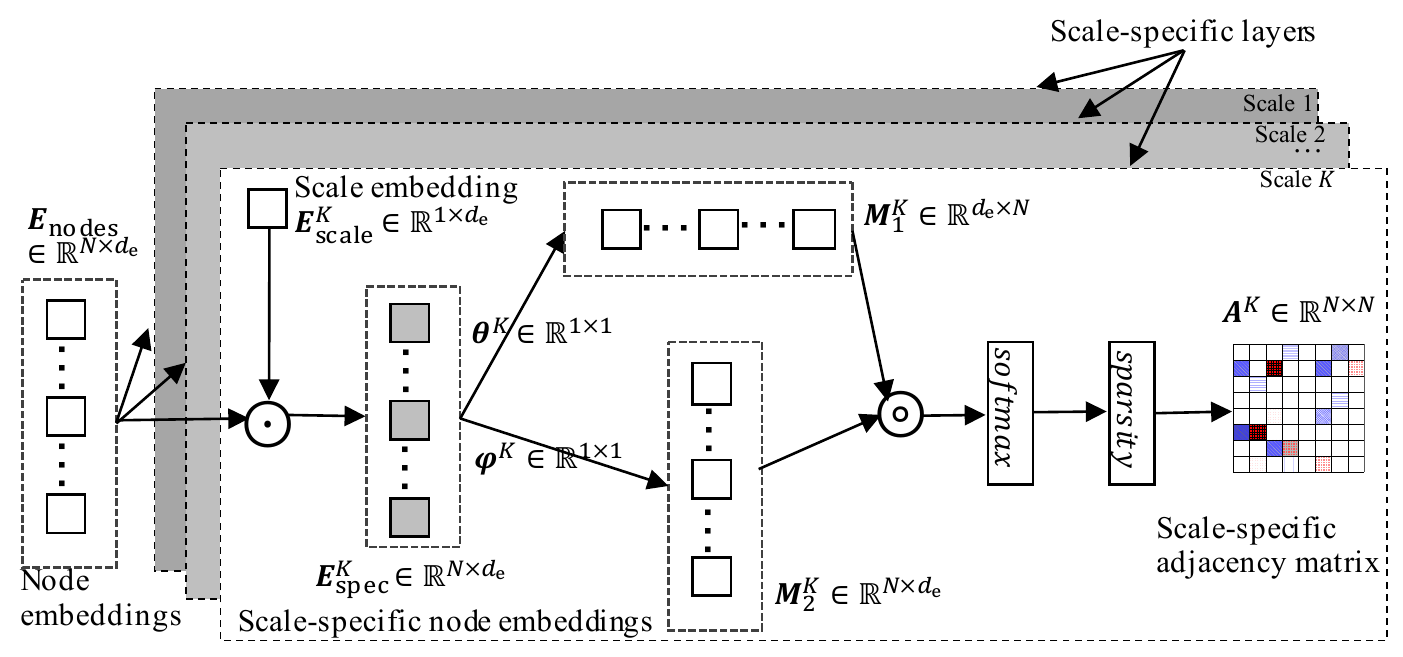}
\caption{The detailed architecture of the adaptive graph learning module.}
\label{Figure_3}
\end{figure}

By such a design, the number of parameters is limited, while $\boldsymbol{E}_{\text{spec}}^k$ contains both the shared node information and the scale-specific information. Then, similar to calculating the node proximities by a similarity function, we calculate pairwise node similarities as follows:
\begin{equation}
\label{8}
\begin{aligned}
&\boldsymbol{M}_1^{k}=[tanh(\boldsymbol{E}^{k}_{\text{spec}}\theta^{k})]^T,\\
&\boldsymbol{M}_2^{k}=tanh(\boldsymbol{E}^{k}_{\text{spec}}\varphi^{k}),\\
&\boldsymbol{A}_{\text{full}}^k=ReLU(\boldsymbol{M}_1^{k}\boldsymbol{M}_2^{k}-(\boldsymbol{M}_2^{k})^T(\boldsymbol{M}_1^{k})^T),
\end{aligned}
\end{equation}
where $\theta^{k}\in\mathbb{R}^{1\times{1}}$ and $\varphi^{k}\in\mathbb{R}^{1\times{1}}$ are learnable parameters to obtain the receiver and sender features of nodes from $\boldsymbol{E}_{\text{spec}}^k$\cite{ref15,ref33}, i.e., $\boldsymbol{M}_1^{k}$ and $\boldsymbol{M}_2^{k}$, respectively. The activation function $tanh$ is used to normalize the input values to $[-1,1]$. The values of $\boldsymbol{A}_{\mathrm{full}}^{k}\in\mathbb{R}^{N\times{N}}$ are then normalized to $[0, 1]$ through the activation function $ReLU$, which are used as the soft edges among the nodes. To reduce the computation cost of the graph convolution, reduce the impact of noise, and make the model more robust, we introduce a strategy to make $\boldsymbol{A}_{\mathrm{full}}^{k}$ sparse:

\begin{equation}
\label{9}
\boldsymbol{A}^{k}=Sparse\left(Softmax(\boldsymbol{A}_{\mathrm{full}}^{k})\right),
\end{equation}
where $\boldsymbol{A}^{k}\in\mathbb{R}^{N\times{N}}$ is the final adjacent matrix of the $k^\text{th}$ layer, $Softmax$ function is used to achieve normalization, and $Sparse$ function is defined as:
\begin{equation}
\label{10}
\boldsymbol{A}_{i j}^{k}= \begin{cases}\boldsymbol{A}_{i j}^{k}, & \boldsymbol{A}_{i j}^{k} \in {TopK}(\boldsymbol{A}_{i *}^{k}, \tau) \\ 0, & \boldsymbol{A}_{i j}^{k} \notin {TopK}(\boldsymbol{A}_{i *}^{k}, \tau)\end{cases},
\end{equation}
where $\tau$ is the threshold of $TopK$ function and denotes the max number of neighbors of a node. The overall architecture of the AGL module is shown in Fig. \ref{Figure_3}. Finally, we can obtain the scale-specific adjacent matrices $\{\boldsymbol{A}^{1}, \ldots, \boldsymbol{A}^{k}, \ldots, \boldsymbol{A}^{K}\}$.
\subsection{Multi-Scale Temporal Graph Neural Network}
Given the multi-scale feature representations  $\{\boldsymbol{X}^{1}, \ldots, \boldsymbol{X}^{k}, \ldots, \boldsymbol{X}^{K}\}$ generated from the multi-scale pyramid network, and the scale-specific adjacent matrices  $\{\boldsymbol{A}^{1}, \ldots, \boldsymbol{A}^{k}, \ldots, \boldsymbol{A}^{K}\}$ generated from the AGL module, a multi-scale temporal graph neural network (MTG) is proposed to capture scale-specific temporal patterns across time steps and variables.

Existing works \cite{ref14, ref18} integrate the GRU and the GNN, which replaces the MLP in the GRU with the GNN to learn inter-variable dependencies. However, the RNN-based solutions often suffer from the gradient vanishing and exploding problems, and adopt the step-by-step strategy for recurrent layers, which makes the model training inefficient, especially when the time series is long enough \cite{ref32}. Temporal convolutional networks (TCNs) have shown superiority on modeling temporal patterns. Thus, we propose a solution that combines the GNNs and temporal convolution layers, i.e., replacing the GRU with temporal convolution layers.

Specifically, the MTG consists of $K$ temporal graph neural networks, each of which combines the TCNs and the GNN to capture scale-specific temporal patterns. For the $k^{\text{th}}$ scale, we first split $\boldsymbol{X}^{k}$ at time dimension and obtain $\{\boldsymbol{x}_{1}^{k}, \ldots, \boldsymbol{x}_{t}^{k}, \ldots, \boldsymbol{x}_{\frac{T}{2^{k-1}}}^{k}\}(\boldsymbol{x}_{t}^{k} \in \mathbb{R}^{N \times c^{k}})$. Similar with \cite{ref15,ref33}, we introduce $\boldsymbol{A}^{k}$ and the transpose of $\boldsymbol{A}^{k}$ (i.e., $(\boldsymbol{A}^{k})^T$ ), and exploit two GNNs to capture both incoming information and outgoing information. Then, the results of GNNs are added:
\begin{equation}
\label{11}
\widetilde{\boldsymbol{h}}_{t}^{k}=GNN_{\mathrm{in}}^{k}(\boldsymbol{x}_{t}^{k}, \boldsymbol{A}^{k}, \boldsymbol{W}_{\mathrm{in}}^{k})+GNN_{\mathrm{out}}^{k}(\boldsymbol{x}_{t}^{k},(\boldsymbol{A}^{k})^{T}, \boldsymbol{W}_{\mathrm{out}}^{k}),
\end{equation}

\noindent where $\boldsymbol{W}_{*}^k$ denotes the trainable parameters of GNNs in the $k^{\text{th}}$ scale. Then, we can obtain all the outputs $\{\widetilde{\boldsymbol{h}}_{1}^{k}, \ldots, \widetilde{\boldsymbol{h}}_{t}^{k}, \ldots, \widetilde{\boldsymbol{h}}_{\frac{T}{2^{k}}}^{k}\}$, which are fed into a temporal convolution layer to obtain the scale-specific representations $\boldsymbol{h}^k$:
\begin{equation}
\label{12}
\boldsymbol{h}^{k}=TCN^{k}(\left[\widetilde{\boldsymbol{h}}_{1}^{k}, \ldots, \widetilde{\boldsymbol{h}}_{t}^{k}, \ldots, \widetilde{\boldsymbol{h}}_{\frac{T}{2^{k}}}^{k}\right], \boldsymbol{W}_{\mathrm{tcn}}^{k}),
\end{equation}
where $\boldsymbol{W}^{k}_{\text{tcn}}$ denotes the trainable parameters in the $k^\text{th}$ temporal convolution layer.

We can see the advantages of exploiting MTG: 1) it can capture scale-specific temporal patterns across time steps and variables; 2) the graph convolution operator enables the model to explicitly consider the inter-variable dependencies.
\subsection{Scale-Wise Fusion}
All the scale-specific representations $\{\boldsymbol{h}^1,...,\boldsymbol{h}^k,...,\boldsymbol{h}^K\}$ can comprehensively reflect all kinds of temporal patterns, where $\boldsymbol{h}^k\in\mathbb{R}^{N\times{d}_\text{s}}$, and $d_{\text{s}}$ denotes the output dimension of TCNs. To obtain the final multi-scale representation, the intuitive solution is to directly concatenate these scale-specific representations or aggregate these representations by a global pooling layer. However, this solution treats each scale-specific representation equally and ignores the difference in contribution to the final forecasting results. For example, the small scale representations are more important for short-term forecasting, while the large scale representations are more important for long-term forecasting. Thus, we propose a scale-wise fusion module to learn a robust multi-scale representation from these scale-specific representations, which can consider the importance of scale-specific temporal patterns and capture the cross-scale correlations.

\begin{figure}[!t]
\centering
\includegraphics[width=3.5in]{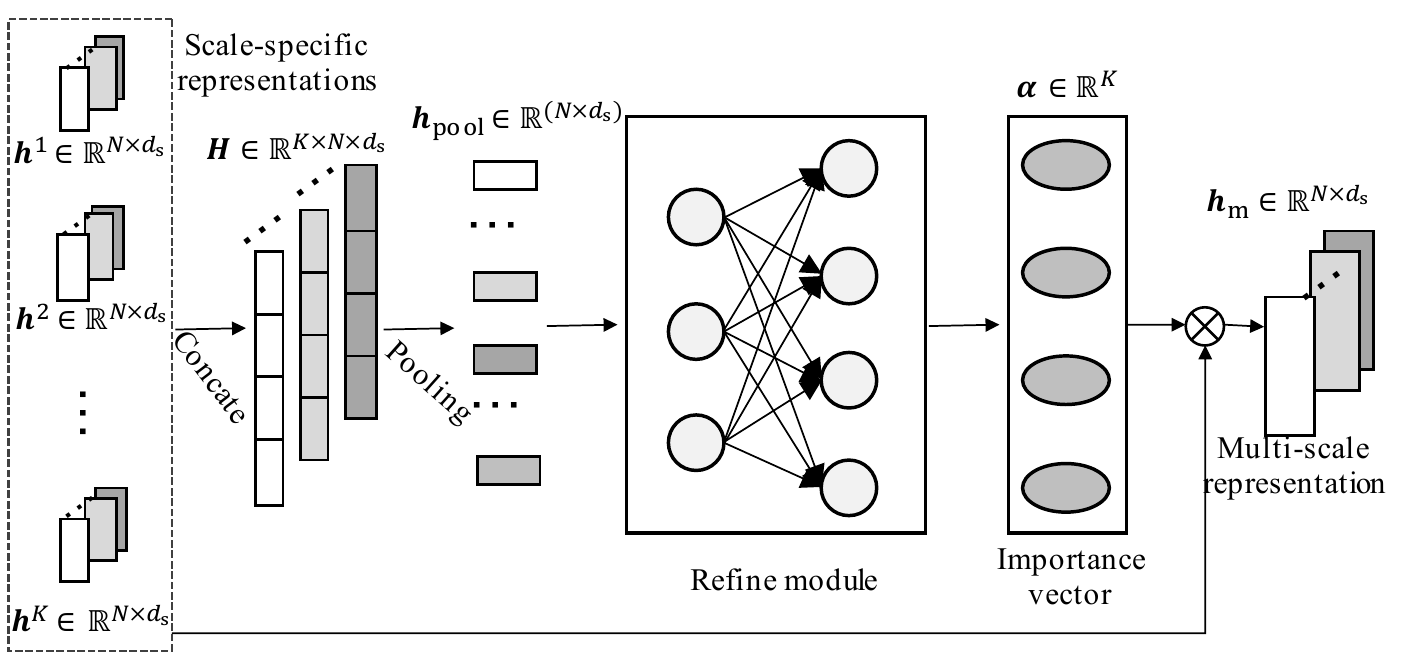}
\caption{The detailed architecture of the scale-wise fusion module.}
\label{Figure_4}
\end{figure}

Fig. \ref{Figure_4} shows the overall architecture of the scale-wise fusion module. Given the scale-specific representations $\{\boldsymbol{h}^1,...,\boldsymbol{h}^k,...,\boldsymbol{h}^K\}$, we first concatenate these representations to obtain the multi-scale matrix $\boldsymbol{H}\in\mathbb{R}^{K\times N\times d_{\text{s}}}$:
\begin{equation}
\label{13}
\boldsymbol{H}={Concat}(\boldsymbol{h}^{1}, \ldots, \boldsymbol{h}^{k}, \ldots, \boldsymbol{h}^{K}),
\end{equation}
where $Concat$ denotes the concatenation operation. Then, we exploit an average pooling layer on the scale dimension:
\begin{equation}
\label{14}
\boldsymbol{h}_{\text {pool }}=\frac{\sum_{k=1}^{K} \boldsymbol{H}^{k}}{K},
\end{equation}
where $\boldsymbol{h}_{\text{pool}}\in\mathbb{R}^{1\times N\times d_\text s}$. Then, we flat $\boldsymbol{h}_{\text{pool}}$ and fed it into a refining module that consists of two full connected layers to compact the fine-grained information across different time scales:
\begin{equation}
\label{15}
\begin{aligned}
&\boldsymbol{\alpha}_{1}={ReLU}\left(\boldsymbol{W}_{1} \boldsymbol{h}_{\text {pool }}+\boldsymbol{b}_{1}\right), \\
&\boldsymbol{\alpha}={Sigmoid}\left(\boldsymbol{W}_{2} \boldsymbol{\alpha}_{1}+\boldsymbol{b}_{2}\right),
\end{aligned}
\end{equation}
where $\boldsymbol{W}_1$ and $\boldsymbol{W}_2$ are weight matrices. $\boldsymbol{b}_1$ and $\boldsymbol{b}_2$ are bias vectors. The $sigmoid$ activation function is used in the second layer. $\boldsymbol{\alpha}\in\mathbb{R}^K$ is defined as the importance score vector that represents the importance of different scale-specific representations. Finally, a weighted aggregation layer is exploited to combine the scale-specific representations:
\begin{equation}
\label{16}
\boldsymbol{h}_{\mathrm{m}}={ReLU}(\sum\nolimits_{k=1}^{K} \boldsymbol{\alpha}[k] \times \boldsymbol{h}^{k}),
\end{equation}
where $\boldsymbol{h}_\text{m}$ is the final multi-scale representation.
\subsection{Output Module \& Objection Function}
The output module includes a convolutional neural network with $1\times d_\text{s}$ kernel size to transform $\boldsymbol{h}_\text{m}\in\mathbb{R}^{N\times d_\text{s}}$ into the desired output dimension, and a followed convolutional neural network with $1\times 1$ kernel size to obtain the predicted values $\boldsymbol{\widehat{x}}\in\mathbb{R}^{N\times 1}$.

The objective function can be formulated as:
\begin{equation}
\label{17}
\mathcal{L}=\frac{1}{\mathcal{T}_{\text {train }}} \sum\nolimits_{i=1}^{\mathcal{T}_{\text {train }}} \sum\nolimits_{j=1}^{N}\left(\widehat{\boldsymbol{x}}_{i, j}-\boldsymbol{x}_{i, j}\right)^{2},
\end{equation}
where $\mathcal{T}_\text{train}$ is the number of training samples, and $N$ is the number of variables. $\boldsymbol{\widehat x}_{i,j}$ and $\boldsymbol{x}_{i,j}$ are the predicted value and ground-truth of the $j^\text{th}$ variable in the $i^\text{th}$ sample, respectively. 
\subsection{Complexity Analysis}
The time complexity of MAGNN consists of the main four modules. For the multi-scale pyramid network, the time complexity of the $k^\text{th}$ scale is $\Theta\left(N \times \frac{T}{2^{k-1}} \times c_{\text{in}} \times c_{\text{out}}\right)$ and the overall time complexity is $\Theta\left(N \times {T} \times c_{\mathrm{in}} \times c_{\mathrm{out}}\right)$, where $N$ is the variable dimension, $T$ is the input sequence length, $c_{\text{in}}$ and $c_{\text{out}}$ are the numbers of input channels and output channels, respectively. Since $c_{\text{in}}$ and $c_{\text{out}}$ are regarded as constants, the time complexity of the multi-scale pyramid network is $\Theta\left(N \times T\right)$. For the AGL module, the time complexity is  $\Theta\left(K \times N \times d_{\mathrm{e}}^{2}+K \times N^{2} \times d_{\mathrm{e}}\right)$, where $K$ is the number of scales and $d_\text{e}$ is the dimension of node or scale embedding. The first half part denotes the point-wise multiplication between node embeddings and scale embeddings. The latter part denotes the pairwise similarity calculation. Since $d_\text{e}$ is regarded as a constant, the time complexity of the AGL module is $\Theta\left(K \times N^2 \right)$. For the MTG module, the time complexity is $\Theta\left(K \left(m\times d_1+ N\times d_\text{in}\times d_\text{s}\right)\right)$, where $m$ denotes the number of edges. $d_\text{in}$ and $d_\text{s}$ denote the input dimension and output dimension, respectively. This result comes from the message passing and information aggregation of GNN. Regarding $d_\text{1}$, $d_\text{in}$, and $d_\text{s}$ as constants, the time complexity of the MTG module is $\Theta\left(K \left(m+N\right)\right)$. For the scale-wise fusion module, the time complexity is $\Theta\left(N\times d_\text{s}\times d_\text{1}+d_\text{1}\times K\right)$, where $d_\text{1}$ is the output dimension of the first full connected layer. Since $d_\text{s}$ and $d_\text{1}$ are regarded as constants, the time complexity of the scale-wise fusion module is $\Theta\left(N+K\right)$.
\section{EXPERIMENTS}
\subsection{Datasets and Settings}
{\bf{Datasets.}} To evaluate the performance of MAGNN, we conduct experiments on six public benchmark datasets: Solar-Energy, Traffic, Electricity, Exchange-Rate, Nasdaq, and METR-LA. Table \ref{tab:table1} gives the summarized dataset statistics, and the details about the six public benchmark datasets are given as follows:
\begin{itemize}
\item{Solar-Energy: This dataset contains the collected solar power from the National Renewable Energy Laboratory, which is sampled every 10 minutes from 137 PV plants in Alabama State in 2007.}
\item{Traffic: This dataset contains the road occupancy rates (between 0 and 1) from the California Department of Transportation, which are hourly aggregated from 862 sensors in San Francisco Bay Area from 2015 to 2016.}
\item{Electricity: This dataset contains the electricity consumptions from the UCI Machine Learning Repository, which are hourly aggregated from 321 clients from 2012 to 2014.}
\item{Exchange-Rate: This dataset contains the exchange rates of eight countries, which are sampled daily from 1990 to 2016.}
\item{Nasdaq: This dataset contains the stock prices of 82 corporations, which are sampled per minute from July 2016 to December 2016.}
\item{METR-LA: This dataset contains the average traffic speeds from Los Angeles County, which are 5-minute aggregated from 207 loop detectors on the highways from March 2012 to June 2012.}
\end{itemize}

Following existing works \cite{ref3,ref11,ref15}, the six datasets are split into the training set (60\%), validation set (20\%), and test set (20\%) in chronological order.
\begin{table}[]
\centering
\caption{Dataset statistics.\label{tab:table1}}
\begin{tabular}{ccccl}
\hline
{Datasets} & {\# Samples} & {\# Variables} & {Sample rate} &  \\ \hline
Solar-Energy      & 52560               & 137                   & 10 minutes           &  \\
Traffic           & 17544               & 862                   & 1 hour               &  \\
Electricity       & 26304               & 321                   & 1 hour               &  \\
Exchange-Rate     & 7588                & 8                     & 1 day                &   \\
Nasdaq           & 40560               & 82                   & 1 minute            &    \\
METR-LA          & 34272               & 207                   & 5 minutes            &    \\ \hline
\end{tabular}
\end{table}

{\bf{Experimental settings.}} MAGNN is implemented in Python with PyTorch 1.7.1 and trained with one GPU (NVIDIA RTX 3090), and the source code is released on GitHub\footnote{https://github.com/shangzongjiang/MAGNN}. For experimental settings, unlike existing works that conduct grid search over all tunable hyper-parameters, we exploit Neural Network Intelligence (NNI)\footnote{ https://nni.readthedocs.io/en/latest/} toolkit to automatically search the best hyper-parameters, which can greatly reduce computation costs. The search spaces of hyper-parameters and the configures of NNI are given in Table \ref{tab:table2}. Following existing works \cite{ref3, ref15}, the input window size $T$ is set to 168. The learning rate is set to 0.001. Adam optimizer is used and all trainable parameters can be optimized through back-propagation. For all datasets, the number of scales is 4. The kernel size of CNNs in the multi-scale pyramid network is set to $1\times 7$, $1\times 6$, and $1\times3$ from the first layer to the final layer of the pyramid network, and the stride size is set to 2 for all CNNs. We set horizon $h=\{3, 6, 12, 24\}$, respectively, which means the forecasting horizons are set from 3 to 24 minutes for Nasdaq dataset, from 15 to 120 minutes for METR-LA dataset, from 30 to 240 minutes for Solar-Energy dataset, from 3 to 24 hours for Traffic and Electricity datasets, and from 3 to 24 days for Exchange-Rate dataset. The larger the forecasting horizon is, the harder the forecasting is. 
\begin{table*}[]
\centering
\caption{Settings of NNI.\label{tab:table2}}
\begin{tabular}{ccccl}
\hline
{}                     & {Parameters}                     & {Choise}                  \\ \hline
\multirow{4}{*}{Search spaces} & Channel size                            & $\{ 8,16,32,64\}$                               \\
                              & Dropout rate                            &  $\{ 0.1,0.5\}$, uniform                      \\    
                              & \# neighbors (Exchange-Rate dataset)         &  $\{ 5,6,7,8\}$                                \\
                              & \# neighbors (the other datasets) &  $\{20,30,40,50\}$                               \\
\multirow{3}{*}{Configures}   & Max trial number                        & 15                               \\
                              & Optimization algorithm                  & Tree-structured Parzen Estimator \\ 
                              & Early stopping strategy                 & Curvefitting                      \\ \hline
\end{tabular}
\end{table*}

{\bf{Evaluation metrics.}} Root Relative Squared Error (RSE) and Empirical Correlation Coefficient (CORR) are exploited as evaluation metrics, which are defined as:
\begin{equation}
\label{18}
\mathrm{RSE}=\frac{\sqrt{\sum_{i=1}^{\mathcal{T}_{\text {test }}} \sum_{j=1}^{N}\left(\boldsymbol{\hat{x}}_{i, j}-\boldsymbol{x}_{i, j}\right)^{2}}}{\sqrt{\sum_{i=1}^{\mathcal{T}_{\text {test }}} \sum_{j=1}^{N}\left(\boldsymbol{x}_{i, j}-\operatorname{mean}(\boldsymbol{x})\right)^{2}}},
\end{equation}

\begin{footnotesize}
\begin{equation}
\label{19}
\text{CORR}=\frac{1}{\mathcal{T}_{\text {test }}} \sum_{j=1}^{N} \frac{\sum_{i=1}^{\mathcal{T}_{\text {test }}}\left(\boldsymbol{x}_{i,j}-\operatorname{mean}\left(\boldsymbol{x}_{*,j}\right)\right)\left(\widehat{\boldsymbol{x}}_{i,j}-\operatorname{mean}\left(\widehat{\boldsymbol{x}}_{*,j}\right)\right)}{\sqrt{\sum_{i=1}^{\mathcal{T}_{\text {test }}}\left(\boldsymbol{x}_{i,j}-\operatorname{mean}\left(\boldsymbol{x}_{*,j}\right)\right)^{2}\left(\widehat{\boldsymbol{x}}_{i,j}-\operatorname{mean}\left(\widehat{\boldsymbol{x}}_{*,j}\right)\right)^{2}}},
\end{equation}
\end{footnotesize}

\noindent where $\mathcal{T}_{\text {test }}$ is the total time steps used for test. For RSE, a lower value is better, while for CORR, a higher value is better.

\subsection{Methods for Comparison}
The methods in our comparative evaluation and the search spaces of their key hyper-parameters are as follows.

{\bf{Conventional methods:}}
\begin{itemize}
\item{AR: It stands for the auto-regressive model. The number of lags is chosen from $\{2^0,2^2,2^4,2^6\}$.}
\item{TRMF \cite{ref6}: It stands for the auto-regressive model using temporal regularized matrix factorization. The hidden dimension size of latent temporal embedding and the regularization coefficient $\lambda$ are chosen from $\{2^2,2^3,...,2^6\}$ and $\{0.1,1,10\}$, respectively.}
\item{GP \cite{ref5}: It stands for the Gaussian process time series model. The RBF kernel bandwidth $\sigma$ and the noise level $\alpha$ are chosen from $\{2^{-10},2^{-8},...,2^{10}\}$.}
\item{VAR-MLP \cite{ref34}: It stands for a hybrid model that combines auto-regressive model (VAR) and multilayer perception (MLP). The size of dense layers is chosen from $\{32,50,100\}$. }
\item{RNN-GRU \cite{ref8}: It stands for the RNN using GRU cell for time series forecasting. The hidden dimension size of RNN layers is chosen from $\{32,50,100\}$.}
\end{itemize}

{\bf{Attentive recurrent methods:}}
\begin{itemize}
\item{LSTNet\cite{ref3}: It introduces the CNNs to capture short-term temporal dependencies, and a recurrent-skip layer to capture long-term periodic patterns. The hidden dimension sizes of recurrent layers, convolutional layers, and recurrent-skip layers are chosen from $\{32,50,100\}$, $\{32,50,100\}$, and $\{20,50,100\}$, respectively.}
\item{MTNet\cite{ref10}: It exploits the memory component and attention mechanism to capture long-term temporal dependencies and periodic patterns. The hidden dimension sizes of GRU and convolutional layers are chosen from $\{32,50,100\}$.}
\item{TPA-LSTM\cite{ref11}: It utilizes an attention mechanism to extract important temporal patterns from different time steps and different variables. The hidden dimension sizes of recurrent and convolutional layers are chosen from $\{32,50,100\}$.}
\end{itemize}

{\bf{MTS modeling with graph learning:}}
\begin{itemize}
\item{Graph WaveNet \cite{ref36}}: It utilizes graph convolutions and dilated 1D convolutions to model spatial-temporal relations. The hidden dimension size of node embedding is chosen from $\{1,3,5,10,15,20,30\}$.
\item{AGCRN \cite{ref14}}: It exploits adaptive graph convolutional recurrent network to infer the inter-variable dependencies. The hidden dimension size of node embedding is chosen from $\{1,3,5,10,15,20,30\}$.
\item{MTHetGNN \cite{ref43}: It utilizes heterogeneous graph embedding module to characterize complex relations among variables. The hidden dimension size of graph convolutional layers is chosen from $\{5,10,15,20,50,100,200\}$.}
\item{MTGNN \cite{ref15}: It uses a graph learning module to learn inter-variable dependencies, and models MTS using GNN and dilated convolution. The number of neighbors for each node is chosen from $\{5,6,7,8,15,30\}$.}
\item{MAGNN: It is our proposed method.}
\end{itemize}

On Solar-Energy, Electricity, Traffic, and Exchange-Rate datasets, most baselines (AR, TRMF, VAR-MLP, GP, RNN-GRU, LSTNet, MTNet, TPA-LSTM, and MTGNN) have been compared in the existing literature \cite{ref3,ref10,ref11,ref15}. Thus, we directly adopt the experimental results in literature. For the results of AGCRN, Graph WaveNet, and MTHetGNN on these four datasets and the results of all baselines on Nasdaq and METR-LA datasets, we use the code released in the original papers and tune the key hyper-parameters according to the validation error by NNI toolkit.
\subsection{Main Results}
Table \ref{tab:table3} and Table \ref{tab:table4} report the evaluation results of all the methods on the six datasets, and the following tendencies can be discerned:

\begin{table*}[]
\centering
\caption{ Results summary (in terms of RSE) of all methods on six datasets. The best results are \textbf{bolded}, and the second best results are \underline{underlined}.\label{tab:table3}}
\resizebox{\linewidth}{!}{
\begin{tabular}{cc|ccccccccccccc}
\hline
\multicolumn{1}{c|}{{\begin{tabular}[c]{@{}c@{}}Datasets\\\end{tabular}}}    & {Horizons}  & {AR}                                                               & {TRMF}                                                             & {VAR-MLP}                                                          & {GP}                                                               & {RNN-GRU}                                                          & {LSTNet}                                                           & {MTNet}                                                            & {TPA-LSTM}                                                         & {AGCRN}                                                            & {Graph WaveNet}                                                    & {MTHetGNN}                                                            & {MTGNN}                                                                     & {MAGNN}                                                                     \\ \hline
\multicolumn{1}{l|}{Solar-Energy}  & \begin{tabular}[c|]{@{}c@{}}3\\ 6\\ 12\\ 24\end{tabular} & \begin{tabular}[c]{@{}c@{}}0.2435\\ 0.3790\\ 0.5911\\ 0.8699\end{tabular} & \begin{tabular}[c]{@{}c@{}}0.2473\\ 0.3470\\ 0.5597\\ 0.9005\end{tabular} & \begin{tabular}[c]{@{}c@{}}0.1922\\ 0.2679\\ 0.4244\\ 0.6841\end{tabular} & \begin{tabular}[c]{@{}c@{}}0.2259\\ 0.3286\\ 0.5200\\ 0.7973\end{tabular} & \begin{tabular}[c]{@{}c@{}}0.1932\\ 0.2628\\ 0.4163\\ 0.4852\end{tabular} & \begin{tabular}[c]{@{}c@{}}0.1843\\ 0.2559\\ 0.3254\\ 0.4643\end{tabular} & \begin{tabular}[c]{@{}c@{}}0.1847\\ 0.2398\\ 0.3251\\ 0.4285\end{tabular} & \begin{tabular}[c]{@{}c@{}}0.1803\\ \underline{0.2347}\\ 0.3234\\ 0.4389\end{tabular} & \begin{tabular}[c]{@{}c@{}}0.1840\\ 0.2432\\ 0.3185\\ \underline{0.4141}\end{tabular} & \begin{tabular}[c]{@{}c@{}}\underline{0.1773}\\ \textbf{0.2279}\\ \underline{0.3068}\\ 0.4206\end{tabular} & \begin{tabular}[c]{@{}c@{}}0.1838\\ 0.2600\\ 0.3169\\ 0.4231\end{tabular} & \begin{tabular}[c]{@{}c@{}}0.1778\\ 0.2348\\ {0.3109}\\ 0.4270\end{tabular}          & {\begin{tabular}[c]{@{}c@{}}\textbf{0.1771}\\ {0.2361}\\ \textbf{0.3015}\\ \textbf{0.4108}\end{tabular}} \\ \hline
\multicolumn{1}{c|}{Traffic}       & \begin{tabular}[c|]{@{}c@{}}3\\ 6\\ 12\\ 24\end{tabular} & \begin{tabular}[c]{@{}c@{}}0.5991\\ 0.6218\\ 0.6252\\ 0.6300\end{tabular} & \begin{tabular}[c]{@{}c@{}}0.6708\\ 0.6261\\ 0.5956\\ 0.6442\end{tabular} & \begin{tabular}[c]{@{}c@{}}0.5582\\ 0.6579\\ 0.6023\\ 0.6146\end{tabular} & \begin{tabular}[c|]{@{}c@{}}0.6082\\ 0.6772\\ 0.6406\\ 0.5995\end{tabular} & \begin{tabular}[c]{@{}c@{}}0.5358\\ 0.5522\\ 0.5562\\ 0.5633\end{tabular} & \begin{tabular}[c]{@{}c@{}}0.4777\\ 0.4893\\ 0.4950\\ 0.4973\end{tabular} & \begin{tabular}[c]{@{}c@{}}0.4764\\ 0.4855\\ 0.4877\\ 0.5023\end{tabular} & \begin{tabular}[c|]{@{}c@{}}0.4487\\ 0.4658\\ 0.4641\\ 0.4765\end{tabular} & \begin{tabular}[c]{@{}c@{}}0.4379\\ \underline{0.4635}\\ 0.4694\\ 0.4707\end{tabular} & \begin{tabular}[c]{@{}c@{}}0.4484\\ 0.4689\\ 0.4725\\ 0.4741\end{tabular} & \begin{tabular}[c]{@{}c@{}}0.4826\\ 0.5198\\ 0.5147\\ 0.5250\end{tabular} & \begin{tabular}[c|]{@{}c@{}}\underline{0.4162}\\ 0.4754\\ \underline{0.4461}\\ \underline{0.4535}\end{tabular}          & \textbf{\begin{tabular}[c]{@{}c@{}}0.4097\\ 0.4555\\ 0.4423\\ 0.4434\end{tabular}} \\ \hline
\multicolumn{1}{c|}{Electricity}   & \begin{tabular}[c]{@{}c@{}}3\\ 6\\ 12\\ 24\end{tabular} & \begin{tabular}[c]{@{}c@{}}0.0995\\ 0.1035\\ 0.1050\\ 0.1054\end{tabular} & \begin{tabular}[c]{@{}c@{}}0.1802\\ 0.2039\\ 0.2186\\ 0.3656\end{tabular} & \begin{tabular}[c]{@{}c@{}}0.1393\\ 0.1620\\ 0.1557\\ 0.1274\end{tabular} & \begin{tabular}[c]{@{}c@{}}0.1500\\ 0.1907\\ 0.1621\\ 0.1273\end{tabular} & \begin{tabular}[c]{@{}c@{}}0.1102\\ 0.1144\\ 0.1183\\ 0.1295\end{tabular} & \begin{tabular}[c]{@{}c@{}}0.0864\\ 0.0931\\ 0.1007\\ 0.1007\end{tabular} & \begin{tabular}[c]{@{}c@{}}0.0840\\ 0.0901\\ 0.0934\\ 0.0969\end{tabular} & \begin{tabular}[c]{@{}c@{}}0.0823\\ 0.0916\\ 0.0964\\ 0.1006\end{tabular} & \begin{tabular}[c]{@{}c@{}}0.0766\\ 0.0894\\ 0.0921\\ 0.0967\end{tabular} & \begin{tabular}[c]{@{}c@{}}\underline{0.0746}\\ 0.0922\\ \underline{0.0909}\\ \underline{0.0962}\end{tabular} & \begin{tabular}[c]{@{}c@{}}0.0749\\ 0.0892\\ 0.0959\\ 0.0969\end{tabular} & {\begin{tabular}[c]{@{}c@{}}\textbf{0.0745}\\ \underline{0.0878}\\ 0.0916\\ \textbf{0.0953}\end{tabular}} & {\begin{tabular}[c]{@{}c@{}}\textbf{0.0745}\\ \textbf{0.0876}\\ \textbf{0.0908}\\ 0.0963\end{tabular}} \\ \hline
\multicolumn{1}{c|}{Exchange-Rate} & \begin{tabular}[c]{@{}c@{}}3\\ 6\\ 12\\ 24\end{tabular} & \begin{tabular}[c]{@{}c@{}}0.0228\\ 0.0279\\ 0.0353\\ 0.0445\end{tabular} & \begin{tabular}[c]{@{}c@{}}0.0351\\ 0.0875\\ 0.0494\\ 0.0563\end{tabular} & \begin{tabular}[c]{@{}c@{}}0.0265\\ 0.0394\\ 0.0407\\ 0.0578\end{tabular} & \begin{tabular}[c]{@{}c@{}}0.0239\\ 0.0272\\ 0.0394\\ 0.0580\end{tabular}  & \begin{tabular}[c]{@{}c@{}}0.0192\\ 0.0264\\ 0.0408\\ 0.0626\end{tabular} & \begin{tabular}[c]{@{}c@{}}0.0226\\ 0.0280\\ 0.0356\\ 0.0449\end{tabular} & \begin{tabular}[c]{@{}c@{}}0.0212\\ 0.0258\\ 0.0347\\ \textbf{0.0442}\end{tabular} & \begin{tabular}[c]{@{}c@{}}\textbf{0.0174}\\ \textbf{0.0241}\\ \textbf{0.0341}\\ \underline{0.0444}\end{tabular} & \begin{tabular}[c]{@{}c@{}}0.0269\\ 0.0331\\ 0.0374\\ 0.0476\end{tabular} & \begin{tabular}[c]{@{}c@{}}0.0251\\ 0.0300\\ 0.0381\\ 0.0486\end{tabular} & \begin{tabular}[c]{@{}c@{}}0.0198\\ 0.0259\\ 0.0345\\ 0.0451\end{tabular} & \begin{tabular}[c]{@{}c@{}}0.0194\\ 0.0259\\ 0.0349\\ {0.0456}\end{tabular}          & \begin{tabular}[c]{@{}c@{}}\underline{0.0183}\\ \underline{0.0246}\\ \underline{0.0343}\\ 0.0474\end{tabular}          \\ \hline
\multicolumn{1}{c|}{Nasdaq}        & \begin{tabular}[c]{@{}c@{}}3\\ 6\\ 12\\ 24\end{tabular} & \begin{tabular}[c]{@{}c@{}}0.0028\\ 0.0029\\ 0.0031\\ 0.0033\end{tabular} & \begin{tabular}[c]{@{}c@{}}0.0020\\ 0.0023\\ 0.0021\\ 0.0026\end{tabular} & \begin{tabular}[c]{@{}c@{}}0.0021\\ 0.0025\\ 0.0027\\ 0.0028\end{tabular} & \begin{tabular}[c]{@{}c@{}}0.0016\\ 0.0024\\ 0.0022\\ 0.0027\end{tabular} & \begin{tabular}[c]{@{}c@{}}0.0014\\ 0.0019\\ 0.0022\\ 0.0024\end{tabular} & \begin{tabular}[c]{@{}c@{}}0.0018\\ 0.0019\\ 0.0021\\ 0.0025\end{tabular} & \begin{tabular}[c]{@{}c@{}}0.0014\\ 0.0019\\ 0.0022\\ 0.0024\end{tabular} & \begin{tabular}[c]{@{}c@{}}\underline{0.0012}\\ \underline{0.0013}\\ \textbf{0.0018}\\ 0.0024\end{tabular} & \begin{tabular}[c]{@{}c@{}}\underline{0.0012}\\ 0.0018\\ 0.0022\\ \underline{0.0023}\end{tabular} & \begin{tabular}[c]{@{}c@{}}0.0018\\ 0.0022\\ 0.0023\\ 0.0026\end{tabular} & \begin{tabular}[c]{@{}c@{}}0.0016\\ 0.0026\\ \underline{0.0020}\\ 0.0030\end{tabular} & \begin{tabular}[c]{@{}c@{}}0.0015\\ 0.0018\\ \underline{0.0020}\\ 0.0026\end{tabular}          & \textbf{\begin{tabular}[c]{@{}c@{}}0.0010\\ 0.0011\\ 0.0018\\ 0.0020\end{tabular}} \\ \hline
\multicolumn{1}{c|}{METR-LA}       & \begin{tabular}[c]{@{}c@{}}3\\ 6\\ 12\\ 24\end{tabular} & \begin{tabular}[c]{@{}c@{}}0.6153\\ 0.6734\\ 0.7579\\ 0.8706\end{tabular} & \begin{tabular}[c]{@{}c@{}}0.5266\\ 0.7907\\ 0.7896\\ 0.8490\end{tabular} & \begin{tabular}[c]{@{}c@{}}0.4240\\ 0.5468\\ 0.6854\\ 0.8635\end{tabular} & \begin{tabular}[c]{@{}c@{}}0.4273\\ 0.5426\\ 0.6752\\ 0.8230\end{tabular} & \begin{tabular}[c]{@{}c@{}}0.4338\\ 0.6865\\ 0.7183\\ 0.8310\end{tabular} & \begin{tabular}[c]{@{}c@{}}0.4259\\ 0.5384\\ 0.6881\\ 0.8202\end{tabular} & \begin{tabular}[c]{@{}c@{}}0.4274\\ 0.5426\\ 0.6752\\ 0.8230\end{tabular} & \begin{tabular}[c]{@{}c@{}}0.4238\\ \underline{0.5371}\\ 0.6758\\ 0.8292\end{tabular} & \begin{tabular}[c]{@{}c@{}}0.4218\\ 0.5485\\ 0.6744\\ 0.8277\end{tabular} & \begin{tabular}[c]{@{}c@{}}0.4206\\ 0.5420\\ 0.6746\\ \underline{0.8054}\end{tabular} & \begin{tabular}[c]{@{}c@{}}0.4217\\ 0.5448\\ 0.6805\\ 0.8324\end{tabular} & {\begin{tabular}[c]{@{}c@{}}\textbf{0.4201}\\ 0.5422\\ \textbf{0.6642}\\ 0.8146\end{tabular}} & \begin{tabular}[c]{@{}c@{}}\underline{0.4202}\\ \textbf{0.5345}\\ \underline{0.6728}\\ \textbf{0.8044}\end{tabular}          \\ \hline
\end{tabular}
}
\end{table*}

\begin{table*}[]
\centering
\caption{ Results summary (in terms of CORR) of all methods on six datasets. The best results are \textbf{bolded}, and the second best results are \underline{underlined}.\label{tab:table4}}
\resizebox{\linewidth}{!}{
\begin{tabular}{cc|ccccccccccccc}
\hline
\multicolumn{1}{c|}{{\begin{tabular}[c]{@{}c@{}}Datasets\\ \end{tabular}}}   &{Horizons}  & {AR}                                                               & {TRMF}                                                             & {VAR-MLP}                                                          & {GP}                                                               & {RNN-GRU}                                                          & {LSTNet}                                                           & {MTNet}                                                            & {TPA-LSTM}                                                                  & {AGCRN}                                                            & {Graph WaveNet}                                                    & {MTHetGNN}                                                         & {MTGNN}                                                                     & {MAGNN}                                                                     \\ \hline
\multicolumn{1}{l|}{Solar-Energy}  & \begin{tabular}[c]{@{}c@{}}3\\ 6\\ 12\\ 24\end{tabular} & \begin{tabular}[c]{@{}c@{}}0.9710\\ 0.9263\\ 0.8107\\ 0.5314\end{tabular} & \begin{tabular}[c]{@{}c@{}}0.9703\\ 0.9418\\ 0.8475\\ 0.5598\end{tabular} & \begin{tabular}[c]{@{}c@{}}0.9829\\ 0.9655\\ 0.9058\\ 0.7149\end{tabular} & \begin{tabular}[c]{@{}c@{}}0.9751\\ 0.9448\\ 0.8518\\ 0.5971\end{tabular} & \begin{tabular}[c]{@{}c@{}}0.9823\\ 0.9675\\ 0.9150\\ 0.8823\end{tabular} & \begin{tabular}[c]{@{}c@{}}0.9843\\ 0.9690\\ 0.9467\\ 0.8870\end{tabular} & \begin{tabular}[c]{@{}c@{}}0.9840\\ 0.9723\\ 0.9462\\ 0.9013\end{tabular} & \begin{tabular}[c]{@{}c@{}}0.9850\\ \underline{0.9742}\\ 0.9487\\ {0.9081}\end{tabular}          & \begin{tabular}[c]{@{}c@{}}0.9841\\ 0.9708\\ 0.9487\\ \underline{0.9087}\end{tabular} & \begin{tabular}[c]{@{}c@{}}0.9846\\ \textbf{0.9743}\\ \underline{0.9527}\\ {0.9055}\end{tabular} & \begin{tabular}[c]{@{}c@{}}0.9845\\ 0.9681\\ 0.9486\\ 0.9031\end{tabular} & \begin{tabular}[c]{@{}c@{}}\underline{0.9852}\\ 0.9726\\ 0.9509\\ 0.9031\end{tabular}          & {\begin{tabular}[c]{@{}c@{}}\textbf{0.9853}\\ 0.9724\\ \textbf{0.9539}\\ \textbf{0.9097}\end{tabular}} \\ \hline
\multicolumn{1}{c|}{Traffic}       & \begin{tabular}[c]{@{}c@{}}3\\ 6\\ 12\\ 24\end{tabular} & \begin{tabular}[c]{@{}c@{}}0.7752\\ 0.7568\\ 0.7544\\ 0.7519\end{tabular} & \begin{tabular}[c]{@{}c@{}}0.6964\\ 0.7430\\ 0.7748\\ 0.7278\end{tabular} & \begin{tabular}[c]{@{}c@{}}0.8245\\ 0.7695\\ 0.7929\\ 0.7891\end{tabular} & \begin{tabular}[c]{@{}c@{}}0.7831\\ 0.7406\\ 0.7671\\ 0.7909\end{tabular} & \begin{tabular}[c]{@{}c@{}}0.8511\\ 0.8405\\ 0.8345\\ 0.8300\end{tabular} & \begin{tabular}[c]{@{}c@{}}0.8721\\ 0.8690\\ 0.8614\\ 0.8588\end{tabular} & \begin{tabular}[c]{@{}c@{}}0.8728\\ 0.8681\\ 0.8644\\ 0.8570\end{tabular} & \begin{tabular}[c]{@{}c@{}}0.8812\\ \underline{0.8717}\\ 0.8717\\ 0.8629\end{tabular}          & \begin{tabular}[c]{@{}c@{}}0.8850\\ 0.8670\\ 0.8679\\ 0.8664\end{tabular} & \begin{tabular}[c]{@{}c@{}}0.8801\\ 0.8674\\ 0.8646\\ 0.8646\end{tabular} & \begin{tabular}[c]{@{}c@{}}0.8643\\ 0.8452\\ 0.8744\\ 0.8418\end{tabular} & \begin{tabular}[c]{@{}c@{}}\underline{0.8963}\\ 0.8667\\ \underline{0.8794}\\ \underline{0.8810}\end{tabular}          & \textbf{\begin{tabular}[c]{@{}c@{}}0.8992\\ 0.8753\\ 0.8815\\ 0.8813\end{tabular}} \\ \hline
\multicolumn{1}{c|}{Electricity}   & \begin{tabular}[c]{@{}c@{}}3\\ 6\\ 12\\ 24\end{tabular} & \begin{tabular}[c]{@{}c@{}}0.8845\\ 0.8632\\ 0.8591\\ 0.8595\end{tabular} & \begin{tabular}[c]{@{}c@{}}0.8538\\ 0.8424\\ 0.8304\\ 0.7471\end{tabular} & \begin{tabular}[c]{@{}c@{}}0.8708\\ 0.8389\\ 0.8192\\ 0.8679\end{tabular} & \begin{tabular}[c]{@{}c@{}}0.8670\\ 0.8334\\ 0.8394\\ 0.8818\end{tabular} & \begin{tabular}[c]{@{}c@{}}0.8597\\ 0.8623\\ 0.8472\\ 0.8651\end{tabular} & \begin{tabular}[c]{@{}c@{}}0.9283\\ 0.9135\\ 0.9077\\ 0.9119\end{tabular} & \begin{tabular}[c]{@{}c@{}}0.9319\\ 0.9226\\ 0.9165\\ 0.9147\end{tabular} & \begin{tabular}[c]{@{}c@{}}0.9439\\ \textbf{0.9337}\\ 0.9250\\ 0.9133\end{tabular}          & \begin{tabular}[c]{@{}c@{}}0.9408\\ 0.9309\\ 0.9222\\ 0.9183\end{tabular} & \begin{tabular}[c]{@{}c@{}}0.9459\\ 0.9310\\ 0.9267\\ \underline{0.9226}\end{tabular} & \begin{tabular}[c]{@{}c@{}}0.9456\\ 0.9307\\ 0.8783\\ 0.8782\end{tabular} & \begin{tabular}[c]{@{}c@{}}\underline{0.9474}\\ {0.9316}\\ \underline{0.9278}\\ \textbf{0.9234}\end{tabular}          & {\begin{tabular}[c]{@{}c@{}}\textbf{0.9476}\\ \underline{0.9323}\\ \textbf{0.9282}\\ 0.9217\end{tabular}} \\ \hline
\multicolumn{1}{c|}{Exchange-Rate} & \begin{tabular}[c]{@{}c@{}}3\\ 6\\ 12\\ 24\end{tabular} & \begin{tabular}[c]{@{}c@{}}0.9734\\ 0.9656\\ 0.9526\\ 0.9357\end{tabular} & \begin{tabular}[c]{@{}c@{}}0.9142\\ 0.8123\\ 0.8993\\ 0.8678\end{tabular} & \begin{tabular}[c]{@{}c@{}}0.8609\\ 0.8725\\ 0.8280\\ 0.7675\end{tabular} & \begin{tabular}[c]{@{}c@{}}0.8713\\ 0.8193\\ 0.8484\\ 0.8278\end{tabular} & \begin{tabular}[c]{@{}c@{}}\underline{0.9786}\\ \textbf{0.9712}\\ 0.9531\\ 0.9223\end{tabular} & \begin{tabular}[c]{@{}c@{}}0.9735\\ 0.9658\\ 0.9511\\ 0.9354\end{tabular} & \begin{tabular}[c]{@{}c@{}}0.9767\\ 0.9703\\ \underline{0.9561}\\ \textbf{0.9388}\end{tabular} & {\begin{tabular}[c]{@{}c@{}}\textbf{0.9790}\\ \underline{0.9709}\\ \textbf{0.9564}\\ \underline{0.9381}\end{tabular}} & \begin{tabular}[c]{@{}c@{}}0.9717\\ 0.9615\\ 0.9531\\ 0.9334\end{tabular} & \begin{tabular}[c]{@{}c@{}}0.9740\\ 0.9640\\ 0.9510\\ 0.9294\end{tabular} & \begin{tabular}[c]{@{}c@{}}0.9769\\ 0.9701\\ 0.9539\\ 0.9360\end{tabular} & \begin{tabular}[c]{@{}c@{}}\underline{0.9786}\\ {0.9708}\\ 0.9551\\ 0.9372\end{tabular}          & \begin{tabular}[c]{@{}c@{}}0.9778\\ \textbf{0.9712}\\ 0.9557\\ {0.9339}\end{tabular}          \\ \hline
\multicolumn{1}{c|}{Nasdaq}        & \begin{tabular}[c]{@{}c@{}}3\\ 6\\ 12\\ 24\end{tabular} & \begin{tabular}[c]{@{}c@{}}0.5055\\ 0.5316\\ 0.4109\\ 0.4427\end{tabular} & \begin{tabular}[c]{@{}c@{}}0.7768\\ 0.7744\\ 0.7562\\ 0.7234\end{tabular} & \begin{tabular}[c]{@{}c@{}}0.8855\\ 0.8937\\ 0.8970\\ 0.8830\end{tabular} & \begin{tabular}[c]{@{}c@{}}0.8960\\ 0.8940\\ 0.8805\\ 0.8837\end{tabular} & \begin{tabular}[c]{@{}c@{}}0.9566\\ 0.9536\\ 0.9446\\ 0.9371\end{tabular} & \begin{tabular}[c]{@{}c@{}}0.9882\\ 0.9865\\ 0.9827\\ 0.9751\end{tabular} & \begin{tabular}[c]{@{}c@{}}0.9851\\ 0.9840\\ 0.9804\\ \underline{0.9837}\end{tabular} & \begin{tabular}[c]{@{}c@{}}0.9745\\ 0.9707\\ 0.9705\\ 0.9627\end{tabular}          & \begin{tabular}[c]{@{}c@{}}0.9878\\ 0.9877\\ 0.9816\\ 0.9701\end{tabular} & \begin{tabular}[c]{@{}c@{}}\underline{0.9953}\\ \underline{0.9943}\\ 0.9840\\ 0.9834\end{tabular} & \begin{tabular}[c]{@{}c@{}}{0.9919}\\ 0.9897\\ \underline{0.9849}\\ 0.9799\end{tabular} & \begin{tabular}[c]{@{}c@{}}0.9912\\ 0.9876\\ 0.9834\\ 0.9754\end{tabular}          & \textbf{\begin{tabular}[c]{@{}c@{}}0.9975\\ 0.9951\\ 0.9864\\ 0.9846\end{tabular}} \\ \hline
\multicolumn{1}{c|}{METR-LA}       & \begin{tabular}[c]{@{}c@{}}3\\ 6\\ 12\\ 24\end{tabular} & \begin{tabular}[c]{@{}c@{}}0.8317\\ 0.7723\\ 0.6770\\ 0.5326\end{tabular} & \begin{tabular}[c]{@{}c@{}}0.8522\\ 0.6216\\ 0.6172\\ 0.4998\end{tabular} & \begin{tabular}[c]{@{}c@{}}0.8963\\ 0.8281\\ 0.7147\\ 0.5038\end{tabular} & \begin{tabular}[c]{@{}c@{}}0.8342\\ 0.8311\\ 0.7153\\ 0.5366\end{tabular} & \begin{tabular}[c]{@{}c@{}}0.8964\\ 0.8004\\ 0.7031\\ 0.5589\end{tabular} & \begin{tabular}[c]{@{}c@{}}0.8965\\ 0.8219\\ 0.7093\\ 0.5790\end{tabular} & \begin{tabular}[c]{@{}c@{}}0.8963\\ 0.8230\\ 0.7156\\ 0.5798\end{tabular} & \begin{tabular}[c]{@{}c@{}}0.8663\\ 0.7955\\ 0.6898\\ 0.5519\end{tabular}          & \begin{tabular}[c]{@{}c@{}}0.8978\\ 0.8245\\ 0.6859\\ 0.5781\end{tabular} & \begin{tabular}[c]{@{}c@{}}0.8985\\ 0.8292\\ 0.6760\\ 0.5846\end{tabular} & \begin{tabular}[c]{@{}c@{}}0.8972\\ \underline{0.8312}\\ 0.7215\\ \textbf{0.5859}\end{tabular} & {\begin{tabular}[c]{@{}c@{}}\textbf{0.8992}\\ 0.8296\\ \underline{0.7279}\\ \underline{0.5855}\end{tabular}} & \begin{tabular}[c]{@{}c@{}}\underline{0.8987}\\ \textbf{0.8328}\\ \textbf{0.7289}\\ \textbf{0.5859}\end{tabular}          \\ \hline
\end{tabular}
}
\end{table*}

\begin{figure}[!t]
\centering
\subfloat[Traffic dataset]{\includegraphics[width=0.25\textwidth]{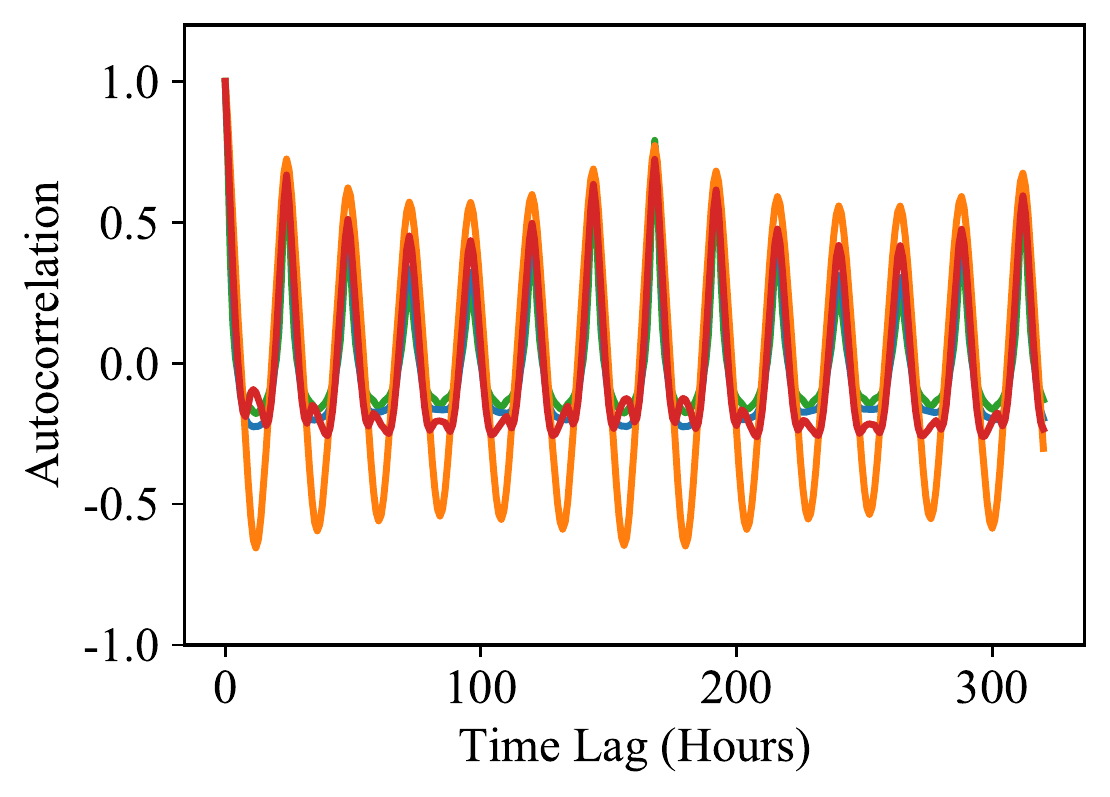}%
\label{fig_first_case}}
\subfloat[Exchange-Rate dataset]{\includegraphics[width=0.25\textwidth]{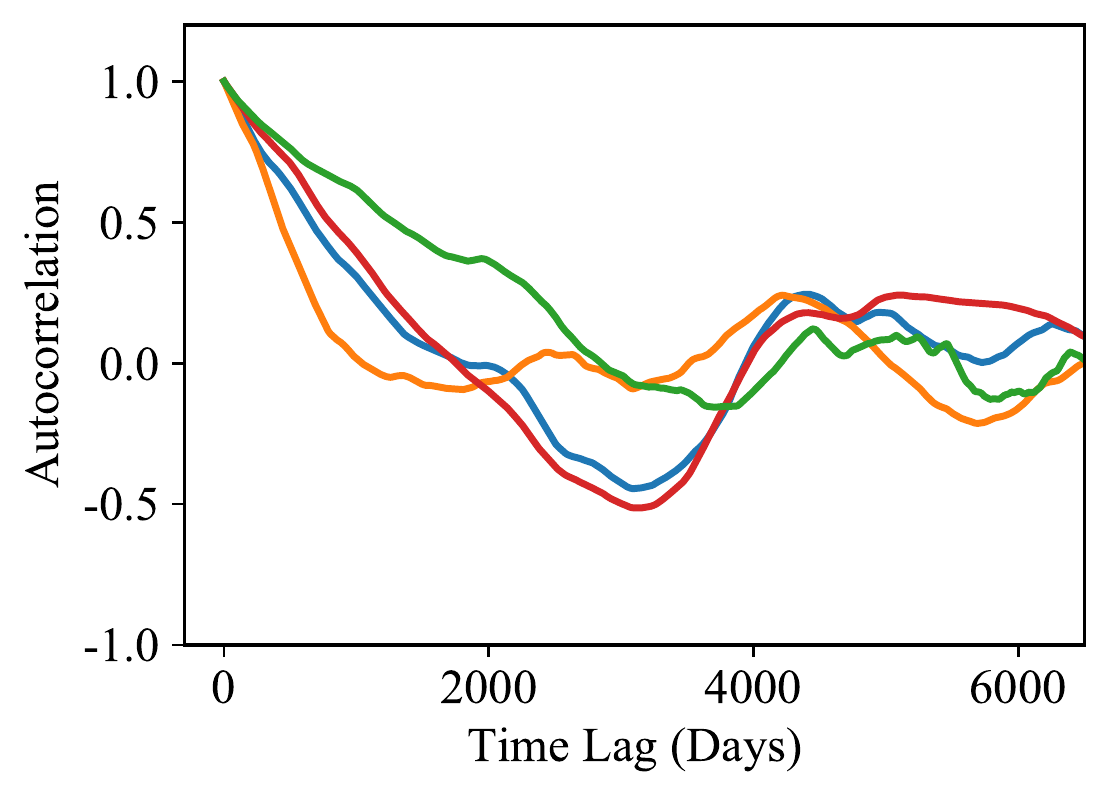}%
\label{fig_third_case}}

\caption{The autocorrelation graphs of four sampled variables.}
\label{Figure_5}
\end{figure}

1) Our method (MAGNN) achieves the state-of-the-art results on these datasets. Particularly, on Traffic and Nasdaq datasets, MAGNN outperforms existing methods on all the horizons and all the metrics. The reason might be that the traffic and stock data are very suitable for our assumption, as there are multi-scale temporal dependencies and complicated inter-variable dependencies. However, on Exchange-Rate dataset, MAGNN obtains slightly worse performance than existing methods. To explore the reasons, Fig. \ref{Figure_5} shows the autocorrelation graphs of sampled variables on Traffic and Exchange-Rate datasets. For Traffic dataset, we can clearly observe the daily and weekly patterns. In contrast, for Exchange-Rate dataset, we can hardly see the multi-scale temporal dependencies. These observations provide empirical guidance for the success of using MAGNN in modeling MTS.

2) Traditional methods (AR, TRMF, and GP) get worse results than deep learning methods, as they cannot capture the non-stationary and non-linear dependencies.

3) Deep learning methods (VAR-MLP, RNN-GRU, LSTNet, MTNet, and TPA-LSTM) do not explicitly model the pairwise inter-variable dependencies. Thus, they get worse performance than AGCRN, Graph WaveNet, MTHetGNN, MTGNN, and MAGNN on most datasets. However, for Exchange-Rate dataset, TPA-LSTM and MTNet outperform most graph-based methods. Specifically, for the RSE evaluation metric, TPA-LSTM achieves the best performance at horizons 3, 6, and 12, and MTNet performs best at horizon 24. One possible explanation for such a phenomenon is that Exchange-Rate dataset only has 7588 samples, leading to the underfitting of graph-based methods that have much more parameters than TPA-LSTM and MTNet.

4) AGCRN, Graph WaveNet, MTHetGNN, and MTGNN are the state-of-the-art methods that use graph learning modules to learn inter-variable dependencies. However, they fail to consider multi-scale inter-variable dependencies and get worse performance than MAGNN in most cases, e.g., MAGNN outperforms MTGNN in 19 out of 24 cases (6 datasets $\times$ 4 horizons) in terms of both RSE and CORR, and exceeds Graph WaveNet in 22 out of 24 cases in terms of both metrics. In contrast, MAGNN learns a temporal representation that can comprehensively reflect both multi-scale temporal patterns and the scale-specific inter-variable dependencies.

\subsection{Effect of Multi-Scale Modeling}
\begin{figure}[!t]
\centering
\subfloat{\includegraphics[width=0.25\textwidth]{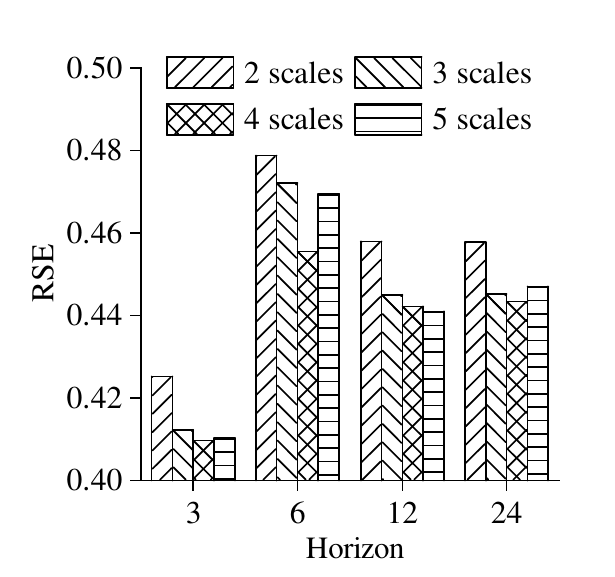}%
\label{fig_first_case}}
\subfloat{\includegraphics[width=0.25\textwidth]{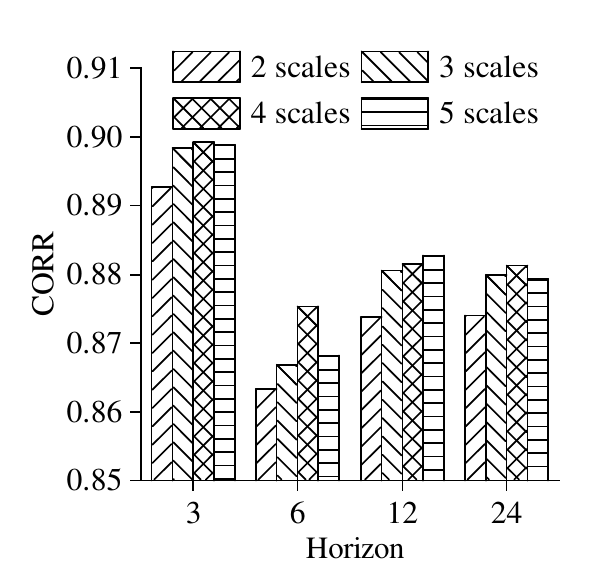}%
\label{fig_second_case}}

\caption{The results of MAGNN under different numbers of scales.}
\label{Figure_6}
\end{figure}

To investigate the effect of multi-scale modeling, we evaluate the performance of MAGNN with different numbers of scales (i.e., 2 scales, 3 scales, 4 scales, and 5 scales). Fig. \ref{Figure_6} shows the results of MAGNN under different numbers of scales on Traffic dataset. We can observe that when the number of scales increases from 2 to 4, the performance of MAGNN is significantly improved. This is because MAGNN can capture more diversified short-term and long-term patterns. When the number of scales increases up to 5, the performance of MAGNN has not improved, which might be because the number of scales is already meet the needs of the task, and excessive parameters are prone to overfitting.

\subsection{Effect of Multi-Scale Feature Extraction }

To investigate the effect of multi-scale feature extraction, we conduct ablation study by carefully designing the following variant.
\begin{itemize}
\item{MAGNN-dila: It extracts multi-scale features by dilated 1D convolutions.}
\end{itemize}

The results of MAGNN-dila and MAGNN on Electricity dataset are shown in Table \ref{tab:table5}. We can see that MAGNN achieves better performance than MAGNN-dila. The reason is that the dilation rates of dilated 1D convolutions may cause the loss of local information, bringing negative effects on modeling short-term dependencies.

\begin{table*}[!t]
\centering
\caption{ The results of different multi-scale feature extraction methods.\label{tab:table5}}
\begin{tabular}{ccccccccc}
\hline
\multicolumn{1}{c}{\multirow{2}{*}{{Methods}}} & \multicolumn{2}{c}{{3}}                                             & \multicolumn{2}{c}{{6}}                                             & \multicolumn{2}{c}{{12}}                                       & \multicolumn{2}{c}{{24}}                                            \\ \cline{2-9} 
\multicolumn{1}{c}{}                                  & \multicolumn{1}{c}{{RSE}}    & \multicolumn{1}{c}{{CORR}}   & \multicolumn{1}{c}{{RSE}}    & \multicolumn{1}{c}{{CORR}}   & \multicolumn{1}{c}{{RSE}} & \multicolumn{1}{c}{{CORR}} & \multicolumn{1}{c}{{RSE}}    & \multicolumn{1}{c}{{CORR}}   \\ \hline
MAGNN-dila                                               & 0.0751                               & 0.9423                               & \textbf{0.0864}                               & \textbf{0.9323}                               & \textbf{0.0899}                   & {0.9237}                    & 0.0969                               & 0.9176                               \\
MAGNN                                            & \textbf{0.0745}                               & \textbf{0.9476}                               & {0.0876}                               & \textbf{0.9323}                               & 0.0908                            & \textbf{0.9282}                             & \textbf{0.0963}                               & \textbf{0.9217}                               \\  \hline
\multicolumn{1}{l}{}                                    & \multicolumn{1}{l}{}                 & \multicolumn{1}{l}{}                 & \multicolumn{1}{l}{}                 & \multicolumn{1}{l}{}                 & \multicolumn{1}{l}{}              & \multicolumn{1}{l}{}               & \multicolumn{1}{l}{}                 & \multicolumn{1}{l}{}                
\end{tabular}
\end{table*}

\subsection{Effect of the Parallel CNNs in the Multi-Scale Pyramid Network}
To demonstrate the effect of the parallel CNNs in the multi-scale pyramid network, we conduct ablation study by carefully designing the following variant.

\begin{itemize}
\item{MAGNN-w/o parallel CNNs: It removes the convolutional neural network with kernel size $1\times 1$ and a $1\times 2$ pooling layer from pyramid layers.}
\end{itemize}

The results presented in Table \ref{tab:table8} show that MAGNN achieves the best performance in all cases on Traffic dataset, indicating the effectiveness of the parallel CNNs. The possible reason for these results is that using the parallel CNNs could make the extracted multi-scale features more stable.

\begin{table*}[!t]
\centering
\caption{ The results of different multi-scale pyramid networks.\label{tab:table8}}
\begin{tabular}{ccccccccc}
\hline
{Methods}        & \multicolumn{2}{c}{{3}}                        & \multicolumn{2}{c}{{6}}                        & \multicolumn{2}{c}{{12}}                       & \multicolumn{2}{c}{{24}}                       \\ \cline{2-9}
{}               & \multicolumn{1}{c}{{RSE}}    & {CORR}   & \multicolumn{1}{c}{{RSE}}    & {CORR}   & \multicolumn{1}{c}{{RSE}}    & {CORR}   & \multicolumn{1}{c}{{RSE}}    & {CORR}   \\ \hline
MAGNN-w/o parallel CNNs & \multicolumn{1}{c}{0.4407}          &{0.8833}          & \multicolumn{1}{c}{0.4567}          & 0.8737          & \multicolumn{1}{c}{0.4688}          & 0.8693          & \multicolumn{1}{c}{0.4739}          & 0.8644          \\ 
MAGNN                   & \multicolumn{1}{c}{\textbf{0.4097}} &\textbf{0.8992} & \multicolumn{1}{c}{\textbf{0.4555}} & \textbf{0.8753} & \multicolumn{1}{c}{\textbf{0.4423}} & \textbf{0.8815} & \multicolumn{1}{c}{\textbf{0.4434}} & \textbf{0.8813} \\ \hline
\end{tabular}
\end{table*}

\subsection{Effect of Adaptive Graph Learning}
To demonstrate the effect of adaptive graph learning, we conduct ablation study by carefully designing the following four variants.

\begin{itemize}
\item{MAGNN-dy: For the $k^\text{th}$ scale at time step $t$, the graph learning module takes the dynamic feature representation $\boldsymbol{x}_t^k$ as input rather than the static node embeddings. Thus, the graphs are different at different time steps.}
\item{MAGNN-full: It removes the sparsity strategy and obtains the multi-scale full-connected adjacent matrices.}
\item{MAGNN-one: It only learns one shared adjacent matrix to describe the inter-variable dependencies of multi-scale feature representations.}
\item{MAGNN-sym: It uses the symmetric adjacency matrix obtained by $\boldsymbol{M}_1^{k}$ and its transpose with one GNN rather than the asymmetric adjacency matrix $\boldsymbol{A}^{k}$ with two GNNs.}
\end{itemize}

The results of these methods on Solar-Energy, Electricity, and Exchange-Rate datasets are shown in Table \ref{tab:table6}, and the following tendencies can be discerned:

1) MAGNN achieves the best performance in most cases, which indicates the superiority of our learned scale-specific adjacency matrices. Specifically, MAGNN performs better than MAGNN-full, MAGNN-one, and MAGNN-sym, showing the effectiveness of the sparsity strategy, the multiple scale-specific graphs, and the asymmetric adjacency matrix, respectively.

2) MAGNN-dy shows competitive performance on Solar-Energy and Exchange-Rate datasets. The results imply the potential of learning dynamic adjacency matrices to model time-varying inter-variate dependencies for forecasting. However, the dramatic fluctuation of adjacency matrices makes MAGNN-dy difficult to maintain stable and excellent performance for all horizons on all datasets. 

\begin{table*}[!t]
\centering
\caption{The results of different graph learning methods.\label{tab:table6}}
\begin{tabular}{cc|cccc|cccc|cccc}
\hline
\multicolumn{2}{c|}{\multirow{2}{*}{{Methods}}}                                               & \multicolumn{4}{c|}{{Solar-Energy}}                                                                                                                                                                                                                                                                                            & \multicolumn{4}{c|}{{Electricity}}                                                                                                                                                                                                                                                                                             & \multicolumn{4}{c}{{Exchange-Rate}}                                                                                                                                                                                                                                                                                                                 \\ \cline{3-14} 
\multicolumn{2}{c|}{}                                                                       & \multicolumn{1}{c}{{3}}                                                             & \multicolumn{1}{c}{{6}}                                                             & \multicolumn{1}{c}{{12}}                                                            & {24}                                                            & \multicolumn{1}{c}{{3}}                                                             & \multicolumn{1}{c}{{6}}                                                             & \multicolumn{1}{c}{{12}}                                                            & {24}                                                            & \multicolumn{1}{c}{{3}}                                                             & \multicolumn{1}{c}{{6}}                                                             & \multicolumn{1}{c}{{12}}                                                            & \multicolumn{1}{c}{{24}}                                                            \\ \hline
\multicolumn{1}{c}{MAGNN-dy}   & \begin{tabular}[c]{@{}l@{}}RSE\\     CORR\end{tabular} & \multicolumn{1}{l}{\begin{tabular}[c]{@{}l@{}}\textbf{0.1766}\\    \textbf{0.9854}\end{tabular}} & \multicolumn{1}{l}{\begin{tabular}[c]{@{}l@{}}0.2372\\    0.9721\end{tabular}} & \multicolumn{1}{l}{\begin{tabular}[c]{@{}l@{}}0.3138\\     0.9504\end{tabular}} & \begin{tabular}[c]{@{}l@{}}\textbf{0.4086}\\     \textbf{0.9109}\end{tabular} & \multicolumn{1}{l}{\begin{tabular}[c]{@{}l@{}}0.0766\\    0.9401\end{tabular}} & \multicolumn{1}{l}{\begin{tabular}[c]{@{}l@{}}0.0864\\     0.9310\end{tabular}} & \multicolumn{1}{l}{\begin{tabular}[c]{@{}l@{}}0.0939\\     0.9191\end{tabular}} & \begin{tabular}[c]{@{}l@{}}0.0987\\     0.9166\end{tabular} & \multicolumn{1}{l}{\begin{tabular}[c]{@{}l@{}}0.0258\\     0.9719\end{tabular}} & \multicolumn{1}{l}{\begin{tabular}[c]{@{}l@{}}0.0310\\     0.9647\end{tabular}} & \multicolumn{1}{l}{\begin{tabular}[c]{@{}l@{}}0.0373\\     0.9534\end{tabular}} & \multicolumn{1}{l}{\begin{tabular}[c]{@{}l@{}}{0.0460}\\     \textbf{0.9373}\end{tabular}} \\ \hline
\multicolumn{1}{l}{MAGNN-full} & \begin{tabular}[c]{@{}l@{}}RSE\\     CORR\end{tabular} & \multicolumn{1}{l}{\begin{tabular}[c]{@{}l@{}}0.1772\\     0.9853\end{tabular}} & \multicolumn{1}{l}{\begin{tabular}[c]{@{}l@{}}0.2398\\     0.9716\end{tabular}} & \multicolumn{1}{l}{\begin{tabular}[c]{@{}l@{}}0.3068\\     0.9522\end{tabular}} & \begin{tabular}[c]{@{}l@{}}0.4246\\     0.9032\end{tabular} & \multicolumn{1}{l}{\begin{tabular}[c]{@{}l@{}}0.0749\\     0.9441\end{tabular}} & \multicolumn{1}{l}{\begin{tabular}[c]{@{}l@{}}\textbf{0.0854}\\     0.9278\end{tabular}} & \multicolumn{1}{l}{\begin{tabular}[c]{@{}l@{}}0.0909\\     0.9236\end{tabular}} & \begin{tabular}[c]{@{}l@{}}0.0976\\     0.9186\end{tabular} & \multicolumn{1}{l}{\begin{tabular}[c]{@{}l@{}}0.0255\\     0.9727\end{tabular}} & \multicolumn{1}{l}{\begin{tabular}[c]{@{}l@{}}0.0284\\     0.9682\end{tabular}} & \multicolumn{1}{l}{\begin{tabular}[c]{@{}l@{}}0.0389\\     0.9514\end{tabular}} & \multicolumn{1}{l}{\begin{tabular}[c]{@{}l@{}}0.0485\\     0.9369\end{tabular}} \\ \hline
\multicolumn{1}{c}{MAGNN-one}  & \begin{tabular}[c]{@{}l@{}}RSE\\     CORR\end{tabular} & \multicolumn{1}{l}{\begin{tabular}[c]{@{}l@{}}0.1769\\     0.9853\end{tabular}} & \multicolumn{1}{l}{\begin{tabular}[c]{@{}l@{}}0.2377\\     0.9720\end{tabular}} & \multicolumn{1}{l}{\begin{tabular}[c]{@{}l@{}}0.3085\\     0.9516\end{tabular}} & \begin{tabular}[c]{@{}l@{}}0.4257\\     0.9021\end{tabular} & \multicolumn{1}{l}{\begin{tabular}[c]{@{}l@{}}0.0776\\     0.9386\end{tabular}} & \multicolumn{1}{l}{\begin{tabular}[c]{@{}l@{}}0.0873\\     0.9287\end{tabular}} & \multicolumn{1}{l}{\begin{tabular}[c]{@{}l@{}}0.0948\\     0.9190\end{tabular}} & \begin{tabular}[c]{@{}l@{}}0.0981\\     0.9142\end{tabular} & \multicolumn{1}{l}{\begin{tabular}[c]{@{}l@{}}0.0277\\     0.9697\end{tabular}} & \multicolumn{1}{l}{\begin{tabular}[c]{@{}l@{}}0.0305\\     0.9644\end{tabular}} & \multicolumn{1}{l}{\begin{tabular}[c]{@{}l@{}}0.0378\\     0.9527\end{tabular}} & \multicolumn{1}{l}{\begin{tabular}[c]{@{}l@{}}0.0475\\     0.9348\end{tabular}} \\ \hline
\multicolumn{1}{c}{MGNN-sym}   & \begin{tabular}[c]{@{}l@{}}RSE\\     CORR\end{tabular} & \multicolumn{1}{l}{\begin{tabular}[c]{@{}l@{}}0.1819\\     0.9844\end{tabular}} & \multicolumn{1}{l}{\begin{tabular}[c]{@{}l@{}}0.2423\\     0.9711\end{tabular}} & \multicolumn{1}{l}{\begin{tabular}[c]{@{}l@{}}0.3139\\     0.9501\end{tabular}} & \begin{tabular}[c]{@{}l@{}}0.4099\\     0.9105\end{tabular} & \multicolumn{1}{l}{\begin{tabular}[c]{@{}l@{}}0.0773\\     0.9385\end{tabular}} & \multicolumn{1}{l}{\begin{tabular}[c]{@{}l@{}}0.0866\\     0.9296\end{tabular}} & \multicolumn{1}{l}{\begin{tabular}[c]{@{}l@{}}0.0928\\     0.9227\end{tabular}} & \begin{tabular}[c]{@{}l@{}}0.0979\\     0.9183\end{tabular} & \multicolumn{1}{l}{\begin{tabular}[c]{@{}l@{}}0.0226\\     0.9759\end{tabular}} & \multicolumn{1}{l}{\begin{tabular}[c]{@{}l@{}}0.0311\\     0.9651\end{tabular}} & \multicolumn{1}{l}{\begin{tabular}[c]{@{}l@{}}0.0372\\     0.9529\end{tabular}} & \multicolumn{1}{l}{\begin{tabular}[c]{@{}l@{}}\textbf{0.0454}\\     0.9366\end{tabular}} \\ \hline
\multicolumn{1}{c}{MAGNN}      & \begin{tabular}[c]{@{}l@{}}RSE\\     CORR\end{tabular} & \multicolumn{1}{l}{\begin{tabular}[c]{@{}l@{}}0.1771\\     0.9853\end{tabular}} & \multicolumn{1}{l}{\begin{tabular}[c]{@{}l@{}}\textbf{0.2361}\\     \textbf{0.9724}\end{tabular}} & \multicolumn{1}{l}{\begin{tabular}[c]{@{}l@{}}\textbf{0.3015}\\     \textbf{0.9539}\end{tabular}} & \begin{tabular}[c]{@{}l@{}}0.4108\\     0.9097\end{tabular} & \multicolumn{1}{l}{\begin{tabular}[c]{@{}l@{}}\textbf{0.0745}\\     \textbf{0.9476}\end{tabular}} & \multicolumn{1}{l}{\begin{tabular}[c]{@{}l@{}}{0.0876}\\     \textbf{0.9323}\end{tabular}} & \multicolumn{1}{l}{\begin{tabular}[c]{@{}l@{}}\textbf{0.0908}\\     \textbf{0.9282}\end{tabular}} & \begin{tabular}[c]{@{}l@{}}\textbf{0.0963}\\     \textbf{0.9217}\end{tabular} & \multicolumn{1}{l}{\begin{tabular}[c]{@{}l@{}}\textbf{0.0183}\\     \textbf{0.9778}\end{tabular}} & \multicolumn{1}{l}{\begin{tabular}[c]{@{}l@{}}\textbf{0.0246}\\     \textbf{0.9712}\end{tabular}} & \multicolumn{1}{l}{\begin{tabular}[c]{@{}l@{}}\textbf{0.0343}\\     \textbf{0.9557}\end{tabular}} & \multicolumn{1}{l}{\begin{tabular}[c]{@{}l@{}}0.0474\\     0.9339\end{tabular}} \\ \hline
\end{tabular}
\end{table*}

\subsection{Effect of Multi-Scale Temporal Graph Neural Network}
To demonstrate the effect of two GNNs in the multi-scale temporal graph neural network, we conduct ablation study by carefully designing the following variant.
\begin{itemize}
\item{MAGNN-one GNN: It only applies one GNN on the learned asymmetric adjacency matrix.}
\end{itemize}

The results presented in Table \ref{tab:table9} show that MAGNN achieves the best performance in all cases on Solar-Energy, Electricity, and Exchange-Rate datasets, indicating the effectiveness of two GNNs. The possible reason for these results is that using two GNNs could exploit more hidden complementary information than using one GNN.

\begin{table*}[]
\centering
\caption{ The results of different multi-scale temporal graph neural networks.\label{tab:table9}}
\begin{tabular}{cc|cccc|cccc|cccc}
\hline
\multicolumn{2}{c|}{Methods}                                                            & \multicolumn{4}{c|}{Solar-Energy}                                                                                                                                                                                                                                                                                                        & \multicolumn{4}{c|}{Electricity}                                                                                                                                                                                                                                                                                                         & \multicolumn{4}{c}{Exchange-Rate}                                                                                                                                                                                                                                                                                                       \\ \cline{3-14}
\multicolumn{2}{c|}{}                                                                   & \multicolumn{1}{c}{3}                                                                & \multicolumn{1}{c}{6}                                                                & \multicolumn{1}{c}{12}                                                               & 24                                                               & \multicolumn{1}{c}{3}                                                                & \multicolumn{1}{c}{6}                                                                & \multicolumn{1}{c}{12}                                                               & 24                                                               & \multicolumn{1}{c}{3}                                                                & \multicolumn{1}{c}{6}                                                                & \multicolumn{1}{c}{12}                                                               & 24                                                               \\ \hline
\multicolumn{1}{c}{MAGNN-one GNN} & \begin{tabular}[c]{@{}c@{}}RSE\\ CORR\end{tabular} & \multicolumn{1}{c}{\begin{tabular}[c]{@{}c@{}}0.1810\\ 0.9847\end{tabular}}          & \multicolumn{1}{c}{\begin{tabular}[c]{@{}c@{}}0.2461\\ 0.9702\end{tabular}}          & \multicolumn{1}{c}{\begin{tabular}[c]{@{}c@{}}0.3104\\ 0.9515\end{tabular}}          & \begin{tabular}[c]{@{}c@{}}0.4271\\ 0.9034\end{tabular}          & \multicolumn{1}{c}{\begin{tabular}[c]{@{}c@{}}0.0777\\ 0.9432\end{tabular}}          & \multicolumn{1}{c}{\begin{tabular}[c]{@{}c@{}}0.0897\\ 0.9221\end{tabular}}          & \multicolumn{1}{c}{\begin{tabular}[c]{@{}c@{}}0.0925\\ 0.9228\end{tabular}}          & \begin{tabular}[c]{@{}c@{}}0.0976\\ 0.9184\end{tabular}          & \multicolumn{1}{c}{\begin{tabular}[c]{@{}c@{}}0.0256\\ 0.9720\end{tabular}}          & \multicolumn{1}{c}{\begin{tabular}[c]{@{}c@{}}0.0297\\ 0.9665\end{tabular}}          & \multicolumn{1}{c}{\begin{tabular}[c]{@{}c@{}}0.0373\\ 0.9525\end{tabular}}          & \begin{tabular}[c]{@{}c@{}}0.0476\\ 0.9332\end{tabular}          \\ \hline
\multicolumn{1}{c}{MAGNN}         & \begin{tabular}[c]{@{}c@{}}RSE\\ CORR\end{tabular} & \multicolumn{1}{c}{\textbf{\begin{tabular}[c]{@{}c@{}}0.1771\\ 0.9853\end{tabular}}} & \multicolumn{1}{c}{\textbf{\begin{tabular}[c]{@{}c@{}}0.2361\\ 0.9724\end{tabular}}} & \multicolumn{1}{c}{\textbf{\begin{tabular}[c]{@{}c@{}}0.3015\\ 0.9539\end{tabular}}} & \textbf{\begin{tabular}[c]{@{}c@{}}0.4108\\ 0.9097\end{tabular}} & \multicolumn{1}{c}{\textbf{\begin{tabular}[c]{@{}c@{}}0.0745\\ 0.9476\end{tabular}}} & \multicolumn{1}{c}{\textbf{\begin{tabular}[c]{@{}c@{}}0.0876\\ 0.9323\end{tabular}}} & \multicolumn{1}{c}{\textbf{\begin{tabular}[c]{@{}c@{}}0.0908\\ 0.9282\end{tabular}}} & \textbf{\begin{tabular}[c]{@{}c@{}}0.0963\\ 0.9217\end{tabular}} & \multicolumn{1}{c}{\textbf{\begin{tabular}[c]{@{}c@{}}0.0183\\ 0.9778\end{tabular}}} & \multicolumn{1}{c}{\textbf{\begin{tabular}[c]{@{}c@{}}0.0246\\ 0.9712\end{tabular}}} & \multicolumn{1}{c}{\textbf{\begin{tabular}[c]{@{}c@{}}0.0343\\ 0.9557\end{tabular}}} & \textbf{\begin{tabular}[c]{@{}c@{}}0.0474\\ 0.9339\end{tabular}} \\ \hline
\end{tabular}
\end{table*}

\subsection{Effect of Scale-Wise Fusion}
To demonstrate the effect of scale-wise fusion, we conduct ablation study by carefully designing the following three variants.
\begin{itemize}
\item{MAGNN-con: It removes the scale-wise fusion module and directly concatenates these scale-specific representations.}
\item{MAGNN-pooling: It removes the scale-wise fusion module and aggregates these scale-specific representations by a global pooling layer.}
\item{MAGNN-att: It replaces the simple concatenation operation in Eq. 13 with attention-based aggregation.}
\end{itemize}

The results of these methods on Solar-Energy, Electricity, and Exchange-Rate datasets are shown in Table \ref{tab:table7}. We can see that, MAGNN achieves the best performance in most cases. The results imply that MAGNN learns a robust multi-scale representation from these scale-specific representations, as our scale-wise fusion can consider the importance of scale-specific temporal patterns and capture the cross-scale correlations.

\begin{figure}[!t]
\centering
\subfloat[Traffic dataset]{\includegraphics[width=0.26\textwidth]{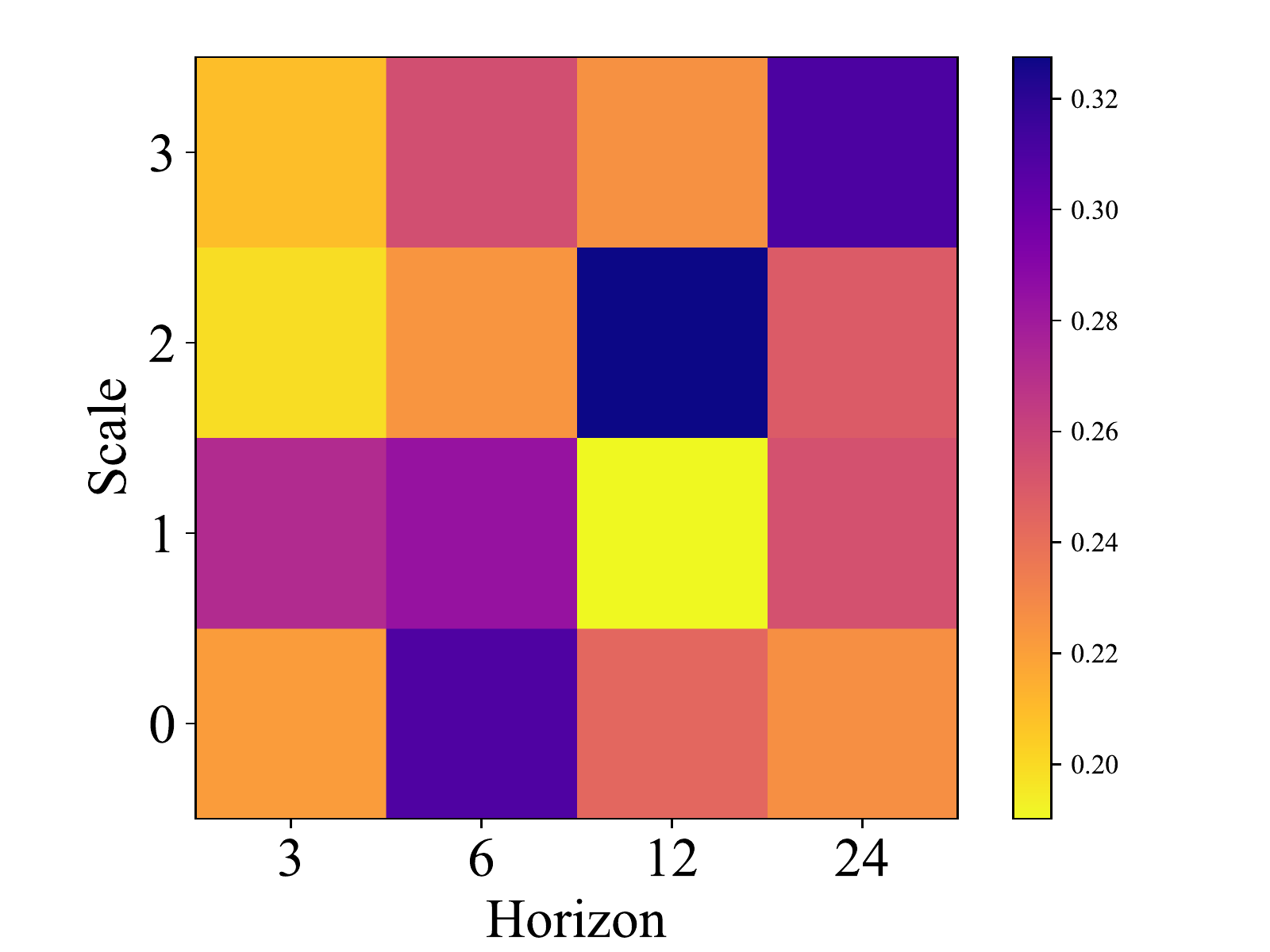}%
\label{fig7_first_case1}}
\subfloat[Solar-Energy dataset]{\includegraphics[width=0.26\textwidth]{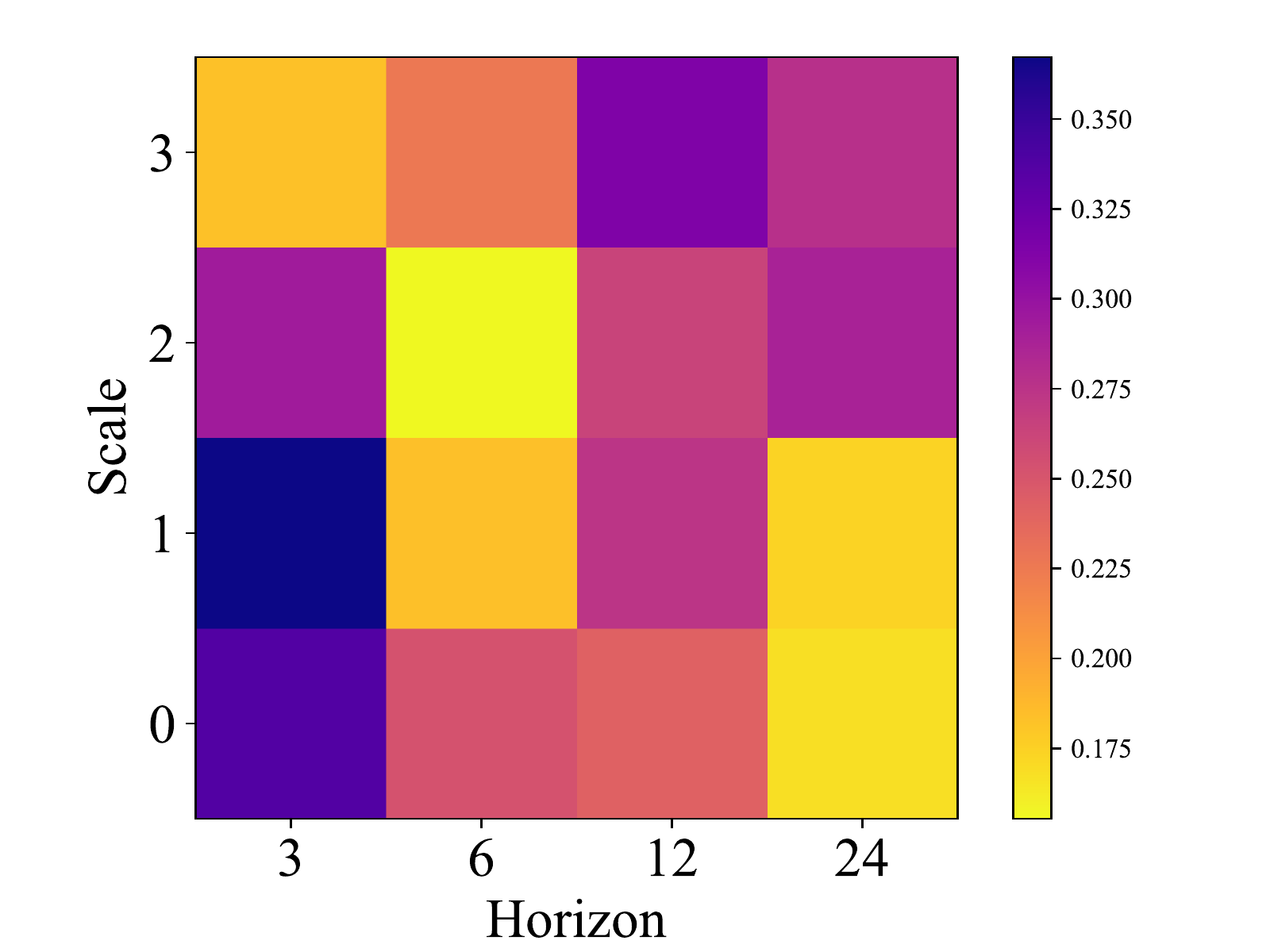}%
\label{fig7_second_case2}}

\caption{Visualization of the weights of MAGNN for Traffic and Solar-Energy datasets. Models are trained with 4 scales (Y-axis) and the forecasting horizon are 3, 6, 12, and 24 (X-axis).}
\label{Figure_7}
\end{figure}

To investigate the effect of different scales, we visualize the weights of temporal representations of different scales for the different forecasting horizons on Traffic and Solar-Energy datasets. The visual results are shown in Fig. \ref{Figure_7}, which indicate that the representations of small scales are more important for short-term forecasting while those of large scales play more essential roles for long-term forecasting.

\begin{table*}[!t]
\centering
\caption{ The results of different fusion methods.\label{tab:table7}}
\begin{tabular}{cc|cccc|cccc|cccc}
\hline
\multicolumn{2}{c|}{Methods}                                                            & \multicolumn{4}{c|}{Solar-Energy}                                                                                                                                                                                                                                                                                                        & \multicolumn{4}{c|}{Electricity}                                                                                                                                                                                                                                                                                                         & \multicolumn{4}{c}{Exchange-Rate}                                                                                                                                                                                                                                                                                                       \\ \cline{3-14}
\multicolumn{2}{c|}{}                                                                   & \multicolumn{1}{c}{3}                                                                & \multicolumn{1}{c}{6}                                                                & \multicolumn{1}{c}{12}                                                               & 24                                                               & \multicolumn{1}{c}{3}                                                                & \multicolumn{1}{c}{6}                                                                & \multicolumn{1}{c}{12}                                                               & 24                                                               & \multicolumn{1}{c}{3}                                                                & \multicolumn{1}{c}{6}                                                                & \multicolumn{1}{c}{12}                                                               & 24                                                               \\ \hline
\multicolumn{1}{c}{MAGNN-con}     & \begin{tabular}[c]{@{}c@{}}RSE\\ CORR\end{tabular} & \multicolumn{1}{c}{\begin{tabular}[c]{@{}c@{}}0.1813\\ 0.9847\end{tabular}}          & \multicolumn{1}{c}{\begin{tabular}[c]{@{}c@{}}0.2452\\ 0.9703\end{tabular}}          & \multicolumn{1}{c}{\begin{tabular}[c]{@{}c@{}}0.3117\\ 0.9513\end{tabular}}          & \begin{tabular}[c]{@{}c@{}}0.4407\\ 0.8954\end{tabular}          & \multicolumn{1}{c}{\begin{tabular}[c]{@{}c@{}}0.0754\\ 0.9428\end{tabular}}          & \multicolumn{1}{c}{{\begin{tabular}[c]{@{}c@{}}\textbf{0.0852}\\ 0.9298\end{tabular}}} & \multicolumn{1}{c}{\begin{tabular}[c]{@{}c@{}}0.0921\\ 0.9221\end{tabular}}          & \begin{tabular}[c]{@{}c@{}}0.0977\\ 0.9168\end{tabular}          & \multicolumn{1}{c}{\begin{tabular}[c]{@{}c@{}}0.0254\\ 0.9726\end{tabular}}          & \multicolumn{1}{c}{\begin{tabular}[c]{@{}c@{}}0.0313\\ 0.9643\end{tabular}}          & \multicolumn{1}{c}{\begin{tabular}[c]{@{}c@{}}0.0391\\ 0.9529\end{tabular}}          & \textbf{\begin{tabular}[c]{@{}c@{}}0.0460\\ 0.9373\end{tabular}} \\ \hline
\multicolumn{1}{c}{MAGNN-pooling} & \begin{tabular}[c]{@{}c@{}}RSE\\ CORR\end{tabular} & \multicolumn{1}{c}{\begin{tabular}[c]{@{}c@{}}0.1855\\ 0.9839\end{tabular}}          & \multicolumn{1}{c}{\begin{tabular}[c]{@{}c@{}}0.2496\\ 0.9692\end{tabular}}          & \multicolumn{1}{c}{\begin{tabular}[c]{@{}c@{}}0.3256\\ 0.9457\end{tabular}}          & \begin{tabular}[c]{@{}c@{}}0.4297\\ 0.9013\end{tabular}          & \multicolumn{1}{c}{\begin{tabular}[c]{@{}c@{}}0.0764\\ 0.9420\end{tabular}}          & \multicolumn{1}{c}{\begin{tabular}[c]{@{}c@{}}0.0867\\ 0.9280\end{tabular}}          & \multicolumn{1}{c}{\begin{tabular}[c]{@{}c@{}}0.0963\\ 0.9193\end{tabular}}          & \begin{tabular}[c]{@{}c@{}}0.0979\\ 0.9168\end{tabular}          & \multicolumn{1}{c}{\begin{tabular}[c]{@{}c@{}}0.0241\\ 0.9745\end{tabular}}          & \multicolumn{1}{c}{\begin{tabular}[c]{@{}c@{}}0.0311\\ 0.9647\end{tabular}}          & \multicolumn{1}{c}{\begin{tabular}[c]{@{}c@{}}0.0370\\ 0.9537\end{tabular}}          & \begin{tabular}[c]{@{}c@{}}0.0475\\ 0.9350\end{tabular}          \\ \hline
\multicolumn{1}{c}{MAGNN-att}     & \begin{tabular}[c]{@{}c@{}}RSE\\ CORR\end{tabular} & \multicolumn{1}{c}{\begin{tabular}[c]{@{}c@{}}0.1817\\ 0.9845\end{tabular}}          & \multicolumn{1}{c}{\begin{tabular}[c]{@{}c@{}}0.2410\\ 0.9714\end{tabular}}          & \multicolumn{1}{c}{\begin{tabular}[c]{@{}c@{}}0.3174\\ 0.9495\end{tabular}}          & \begin{tabular}[c]{@{}c@{}}0.4368\\ 0.8957\end{tabular}          & \multicolumn{1}{c}{\begin{tabular}[c]{@{}c@{}}0.0772\\ 0.9423\end{tabular}}          & \multicolumn{1}{c}{\begin{tabular}[c]{@{}c@{}}0.0868\\ 0.9303\end{tabular}}          & \multicolumn{1}{c}{\begin{tabular}[c]{@{}c@{}}0.0925\\ {0.9231}\end{tabular}}          & \begin{tabular}[c]{@{}c@{}}0.0979\\ 0.9190\end{tabular}          & \multicolumn{1}{c}{\begin{tabular}[c]{@{}c@{}}0.0270\\ 0.9706\end{tabular}}          & \multicolumn{1}{c}{\begin{tabular}[c]{@{}c@{}}0.0304\\ 0.9662\end{tabular}}          & \multicolumn{1}{c}{\begin{tabular}[c]{@{}c@{}}0.0387\\ 0.9527\end{tabular}}          & \begin{tabular}[c]{@{}c@{}}0.0462\\ 0.9356\end{tabular}          \\ \hline
\multicolumn{1}{c}{MAGNN}         & \begin{tabular}[c]{@{}c@{}}RSE\\ CORR\end{tabular} & \multicolumn{1}{c}{\textbf{\begin{tabular}[c]{@{}c@{}}0.1771\\ 0.9853\end{tabular}}} & \multicolumn{1}{c}{\textbf{\begin{tabular}[c]{@{}c@{}}0.2361\\ 0.9724\end{tabular}}} & \multicolumn{1}{c}{\textbf{\begin{tabular}[c]{@{}c@{}}0.3015\\ 0.9539\end{tabular}}} & \textbf{\begin{tabular}[c]{@{}c@{}}0.4108\\ 0.9097\end{tabular}} & \multicolumn{1}{c}{\textbf{\begin{tabular}[c]{@{}c@{}}0.0745\\ 0.9476\end{tabular}}} & \multicolumn{1}{c}{\begin{tabular}[c]{@{}c@{}}0.0876\\ \textbf{0.9323}\end{tabular}}          & \multicolumn{1}{c}{{\begin{tabular}[c]{@{}c@{}}\textbf{0.0908}\\ \textbf{0.9282}\end{tabular}}} & \textbf{\begin{tabular}[c]{@{}c@{}}0.0963\\ 0.9217\end{tabular}} & \multicolumn{1}{c}{\textbf{\begin{tabular}[c]{@{}c@{}}0.0183\\ 0.9778\end{tabular}}} & \multicolumn{1}{c}{\textbf{\begin{tabular}[c]{@{}c@{}}0.0246\\ 0.9712\end{tabular}}} & \multicolumn{1}{c}{\textbf{\begin{tabular}[c]{@{}c@{}}0.0343\\ 0.9557\end{tabular}}} & \begin{tabular}[c]{@{}c@{}}0.0474\\ 0.9339\end{tabular}          \\ \hline
\end{tabular}
\end{table*}

\begin{table*}[]
\centering
\caption{ The computation costs of different methods.\label{tab:table10}}
\begin{tabular}{cccccc}
\hline
{Methods} & {\# Parameters} & {Training time/epoch} & {Total training time} & {RSE}    & {CORR}   \\ \hline
LSTNet           & \textbf{71613}         & \textbf{34.11s}               & \textbf{0.94h}               & 0.4777          & 0.8721          \\ 
TPA-LSTM         & 379051                 & 313.41s                       & 8.71h                        & 0.4487          & 0.8812          \\ 
MTGNN            & 337345                 & 349.57s                       & 4.86h                        & 0.4162          & 0.8963          \\ 
MAGNN            & 163325                 & 111.89s                       & 1.55h                        & \textbf{0.4097} & \textbf{0.8992} \\ \hline
\end{tabular}
\end{table*}
\subsection{Parameter Study}
\begin{figure}[!t]
\centering
\subfloat[Traffic dataset]{\includegraphics[width=0.23\textwidth]{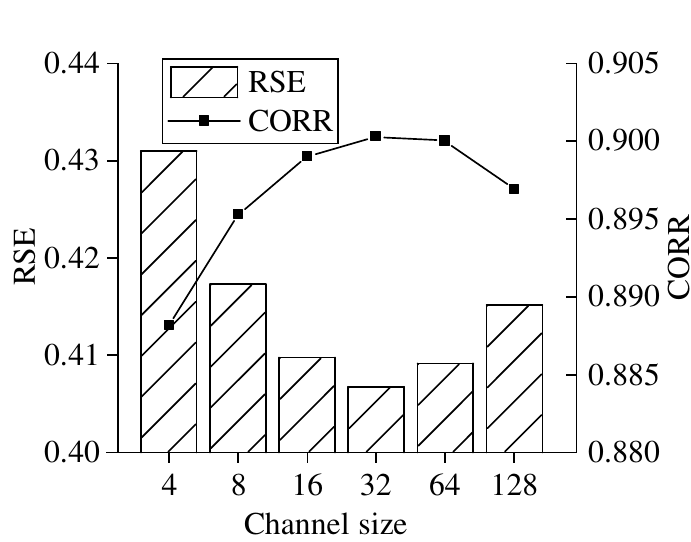}%
\label{fig_first_case1}}
\quad
\subfloat[Solar-Energy dataset]{\includegraphics[width=0.23\textwidth]{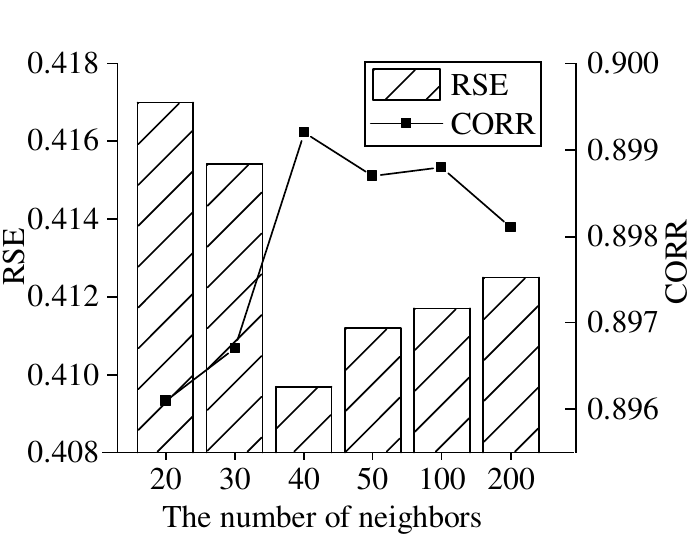}%
\label{fig_second_case2}}

\caption{The effects of hyper-parameters.}
\label{Figure_8}
\end{figure}

We study the two important parameters (i.e., convolutional channel size and the number of neighbors), which could influence the performance of MAGNN. Fig. \ref{fig_first_case1} shows the results of MAGNN on Traffic dataset by varying convolutional channel size from 4 to 128. The best performance can be obtained when convolutional channel size is 32. It might be that a small convolutional channel size limits the expressive ability of MAGNN, and a large convolutional channel size would make the model hard to train. Fig. \ref{fig_second_case2} shows the results of MAGNN on Traffic dataset by varying the number of neighbors from 20 to 200. The best performance can be obtained when the number of neighbors is 40. The reason may be that a small number of neighbors limits the ability to exploit inter-variable dependencies, and a large number of neighbors would introduce noises.

\subsection{Computation Cost}
To evaluate the computation cost, we compare the parameter numbers, training time, and forecasting performances of MAGNN, LSTNet, TPA-LSTM, and MTGNN on Traffic dataset in Table \ref{tab:table10}. LSTNet has least parameter number and runs fastest in these methods. But it gets worst forecasting results. Compared with TPA-LSTM and MTGNN, MAGNN runs fastest and gets best forecasting results. Overall, comprehensively considering the significant forecasting performance improvement and the computation cost, MAGNN demonstrates the superiority over existing methods.

\section{CONCLUSIONS and Future Work}
In this paper, we propose a multi-scale adaptive graph neural network (MAGNN) for MTS forecasting. By exploiting a multi-scale pyramid network to model temporal hierarchy, an adaptive graph learning module to automatically infer inter-variable dependencies, a multi-scale temporal graph neural network to model intra-variable and inter-variable dependencies, and a scale-wise fusion module to promote the collaboration across different time scales, MAGNN outperforms the state-of-the-art methods on six datasets. With the theoretical analysis and experimental verification, we believe that MAGNN can capture multi-scale temporal patterns and complicated inter-variable dependencies for accurate MTS forecasting.

In the future, it is of interest to extend this work in the following three aspects: First, we will design a method to learn dynamic adjacency matrices at different time steps, and introduce a regularizer to constrain the dramatic fluctuation of adjacent matrices. Second, we will design a  neural architecture search framework to automatically capture both inter-variable dependencies and intra-variable dependencies. Third, we will further develop a graph matching-based AGL module by evaluating the structural and semantic similarities of multi-scale graphs, which can reduce time complexity and enhance scalability.
\section*{acknowledgments}
This work was supported by the National Key Research and Development Program of China under Grant 2018YFB0505000 and Alibaba-Zhejiang University Joint Research Institute of Frontier Technologies.

\bibliographystyle{unsrt}
\bibliography{MAGNNV3}
\vspace{-15 mm}
\begin{IEEEbiography}[{\includegraphics[width=1in,height=1.25in,clip,keepaspectratio]{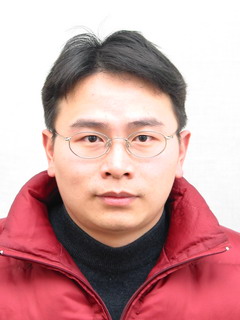}}]
{Ling Chen} received his B.S. and Ph.D. degrees in computer science from Zhejiang University, China, in 1999 and 2004, respectively. He is currently a professor with the College of Computer Science and Technology, Zhejiang University, China. His research interests include ubiquitous computing and data mining.
\end{IEEEbiography}
\vspace{-15 mm}

\begin{IEEEbiography}[{\includegraphics[width=1in,height=1.25in,clip,keepaspectratio]{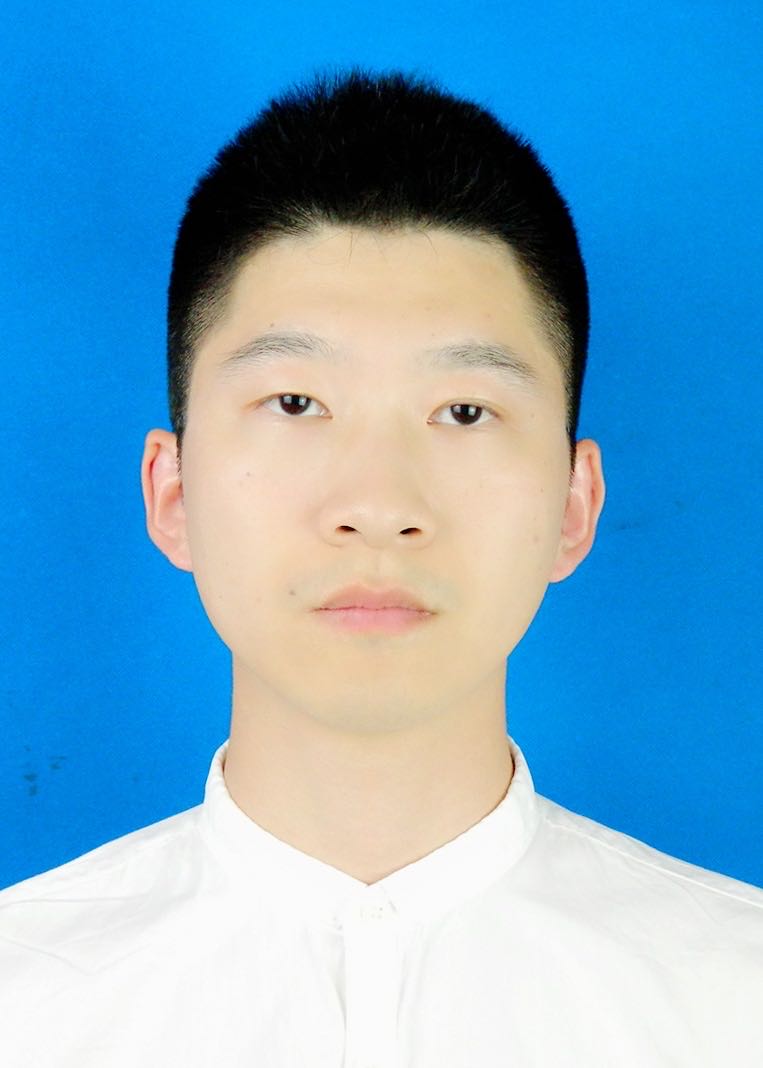}}]
{Donghui Chen} received his Ph.D. degree in computer science from  Zhejiang University, China, in 2021. His research interests include time series representation learning and prediction. 
\end{IEEEbiography}
\vspace{-15 mm}

\begin{IEEEbiography}[{\includegraphics[width=1in,height=1.25in,clip,keepaspectratio]{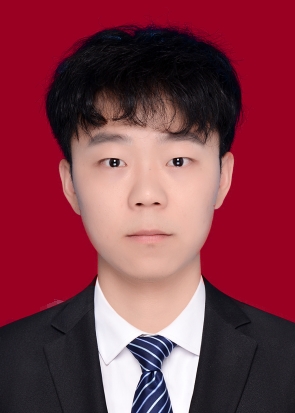}}]
{Zongjiang Shang} received his M.S. degree in electronic information from Northwestern Polytechnical University, China, in 2020. He is currently a Ph.D. student with the College of Computer Science and Technology, Zhejiang University, China. His research interests include time series representation learning and prediction.
\end{IEEEbiography}
\vspace{-15 mm}

\begin{IEEEbiography}[{\includegraphics[width=1in,height=1.25in,clip,keepaspectratio]{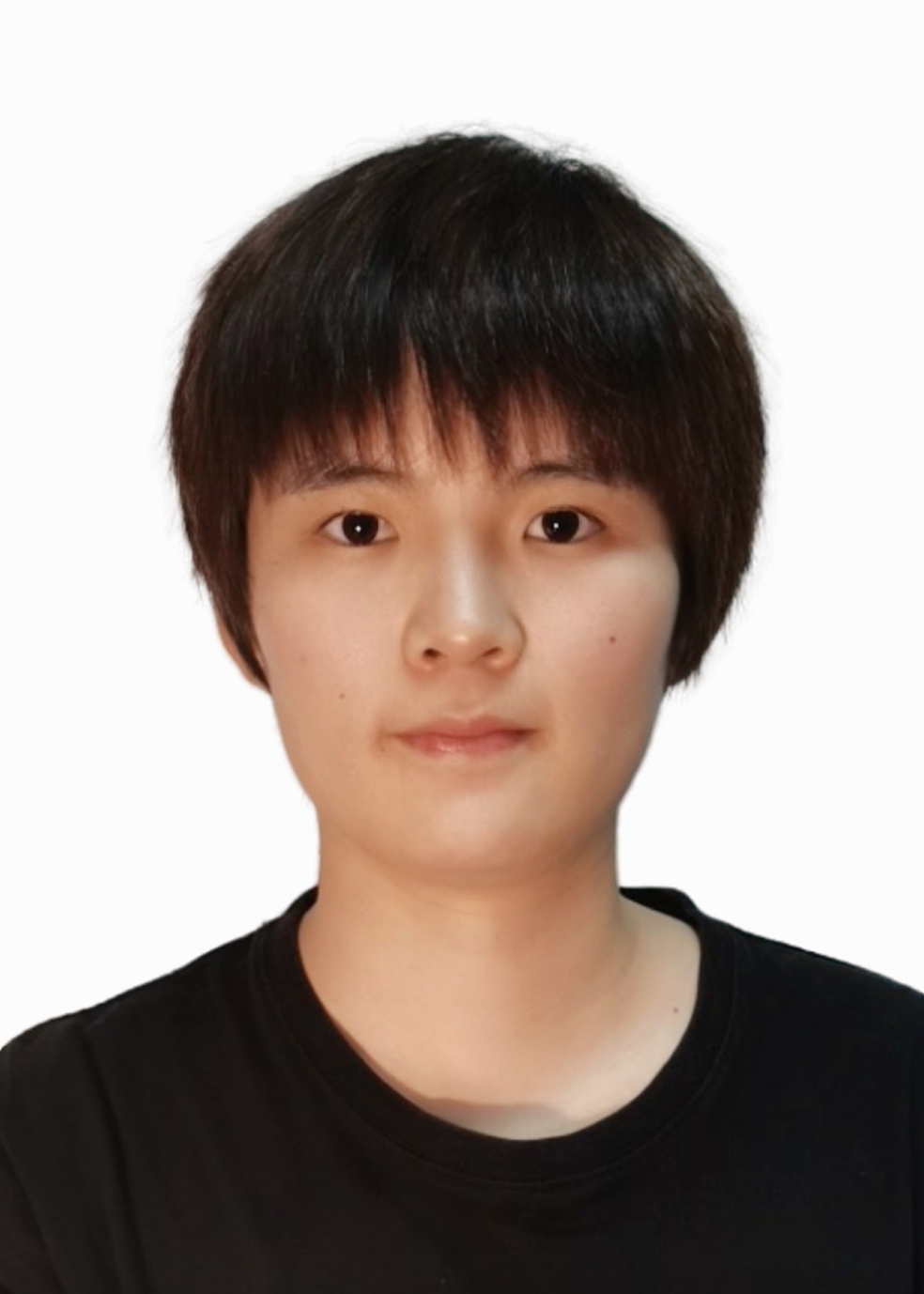}}]
{Binqing Wu} received her B.Eng. degree in computer science from Southwest Jiangtong University, China, in 2020. She is currently a Ph.D. student with the College of Computer Science and Technology, Zhejiang University, China. Her research interests include urban computing and data mining.
\end{IEEEbiography}
\vspace{-15 mm}

\begin{IEEEbiography}[{\includegraphics[width=1in,height=1.25in,clip,keepaspectratio]{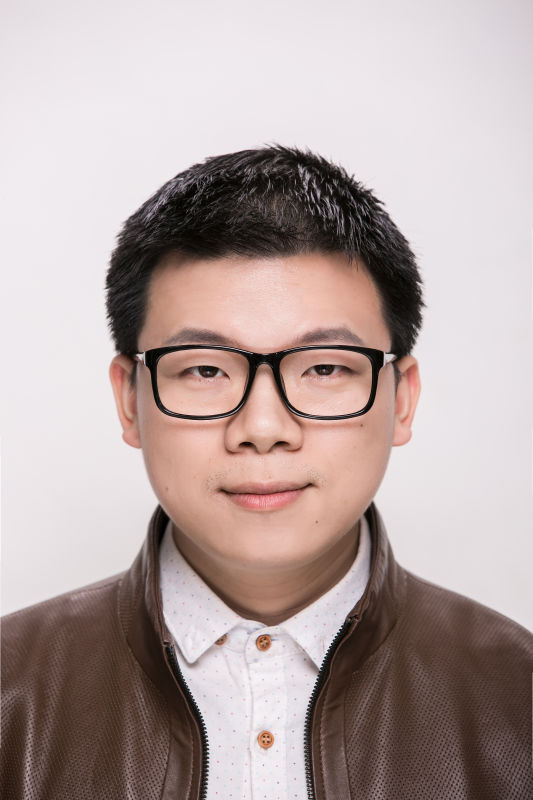}}]
{Cen Zheng} received his M.S. degree in computer science from Shanghai Jiao Tong University, China, in 2011. He is currently a staff engineer in Alibaba Group. His research interests include distributed storage and database.
\end{IEEEbiography}
\vspace{-15 mm}

\begin{IEEEbiography}[{\includegraphics[width=1in,height=1.25in,clip,keepaspectratio]{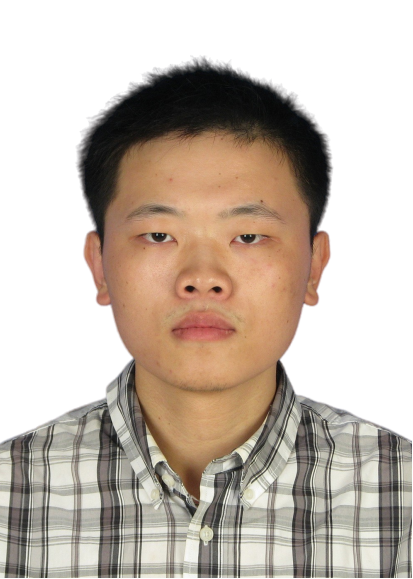}}]
{Bo Wen} received his B.S. degree in computer science from Jishou University, China, in 2012. He is currently a software engineer in Alibaba Group. His research interests include time-series analytics and time-series oriented database.
\end{IEEEbiography}
\vspace{-15 mm}

\begin{IEEEbiography}[{\includegraphics[width=1in,height=1.25in,clip,keepaspectratio]{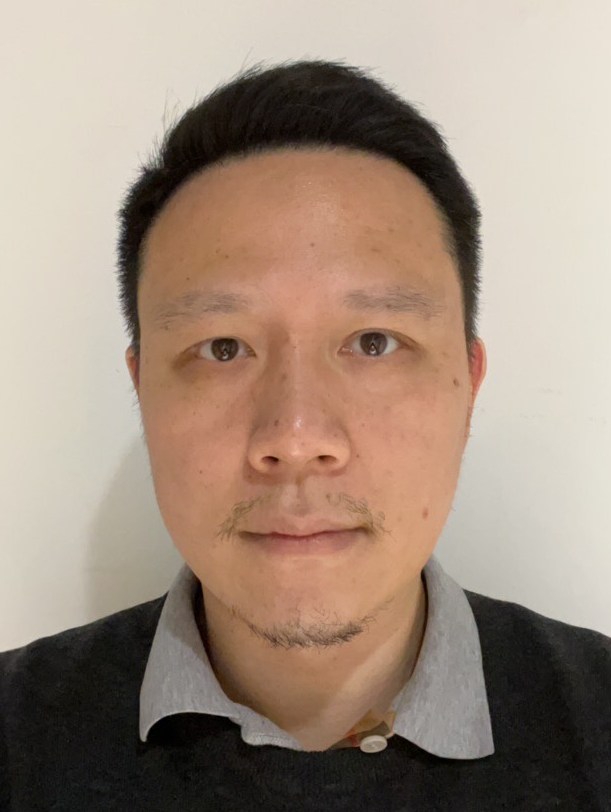}}]
{Wei Zhang} received his Ph.D. from UCSB. He is currently a principal engineer and leads the NoSQL-Database team in Alibaba Group. His research interests include storage, database, and AI.
\end{IEEEbiography}
\vspace{100 mm}

\end{document}